\documentclass[letterpaper,journal]{IEEEtran}
\usepackage{amsmath,amsfonts}
\pdfoutput=1
\usepackage{algorithmic}
\usepackage{array}
\usepackage{textcomp}
\usepackage{stfloats}
\usepackage{url}
\usepackage{verbatim}
\usepackage{graphicx}
\usepackage{tabularx}
\usepackage{xcolor}
\usepackage{lettrine}
\usepackage{graphicx} 
\usepackage{booktabs}  
\usepackage{multirow}  
\usepackage{array}     
\usepackage{threeparttable} 
\usepackage{amsmath}  
\usepackage{float}     
\usepackage{graphicx}
\usepackage{booktabs}
\usepackage{hyperref}
\usepackage{orcidlink}
\usepackage{booktabs}
\usepackage{multirow}
\usepackage{xcolor}
\usepackage{colortbl}

\def\BibTeX{{\rm B\kern-.05em{\sc i\kern-.025em b}\kern-.08em
    T\kern-.1667em\lower.7ex\hbox{E}\kern-.125emX}}
\usepackage{balance}
\begin{document}

\bstctlcite{IEEEexample:BSTcontrol}

\title{\fontsize{22}{25}\selectfont ConsDreamer: Advancing Multi-View Consistency for Zero-Shot Text-to-3D Generation}

\author{Yuan Zhou, Shilong Jin, Litao Hua, Wanjun Lv, Haoran Duan,~\IEEEmembership{Member~IEEE}\\ Jungong Han,~\IEEEmembership{Senior Member~IEEE}

\thanks{
Received 5 April 2025; revised 11 October 2025 and 12 February 2026; accepted 19 April 2026. This work was supported in part by Beijing Natural Science Foundation under Grant L257005.

Yuan Zhou, Shilong Jin and Litao Hua are with the School of Artificial Intelligence, Nanjing University of Information Science and Technology, Jiangsu 210044, China (e-mail: zhouyuan@nuist.edu.cn, shilonnng@gmail.com; wulitao123321@gmail.com).

Jungong Han and Haoran Duan are with the Department of Automation, Tsinghua University, Beijing, China (e-mail: jungonghan77@gmail.com, haoran.duan@ieee.org). The corresponding author: Haoran Duan.

Wanjun Lv is with the Lenovo, Beijing, China (e-mail: lvwj1@lenovo.com).
}
}

\markboth{IEEE Transactions on Image Processing}
{Shell \MakeLowercase{\textit{et al.}}: A Sample Article Using IEEEtran.cls for IEEE Journals}

\IEEEpubid{\begin{minipage}{\textwidth}\ \\[30pt] \centering
		Copyright \copyright 20xx IEEE. Personal use of this material is permitted. 
		However, permission to use this material for any other purposes must \\ be obtained 
		from the IEEE by sending an email to pubs-permissions@ieee.org.
\end{minipage}}

\maketitle
\begin{abstract}
Recent advances in zero-shot text-to-3D generation have revolutionised 3D content creation by enabling direct synthesis from textual descriptions. While state-of-the-art methods leverage 3D Gaussian Splatting with score distillation to enhance multi-view rendering through pre-trained text-to-image~(T2I) models, they suffer from inherent prior view biases in T2I Models. These biases lead to inconsistent 3D generation, particularly manifesting as the multi-face Janus problem, where objects exhibit conflicting features across views. To address this fundamental challenge, we propose \textbf{\textit{ConsDreamer}}, a novel method that mitigates view bias by refining both the conditional and unconditional terms in the score distillation process: (1) a View Disentanglement Module~(VDM) that eliminates viewpoint biases in conditional prompts by decoupling irrelevant view components and injecting precise view control; and (2) a similarity-based partial order loss that enforces geometric consistency in the unconditional term by aligning cosine similarities with azimuth relationships. Extensive experiments demonstrate that ConsDreamer can be seamlessly integrated into various 3D representations and score distillation paradigms, effectively mitigating the multi-face Janus problem. 
\end{abstract}

\begin{IEEEkeywords}
text to 3D generation, multi-face Janus problem, score distillation sampling.
\end{IEEEkeywords}

\section{Introduction}

\lettrine[lines=2]{\textbf{3D}} GENERATION technology plays a crucial role in various fields such as innovative industrial design, game development, and virtual reality. In particular, zero-shot text-to-3D generation~\cite{chen2023fantasia3d-lucid-5}, \cite{hong2022avatarclip-lucid-14}, \cite{lin2023magic3d-lucid-21}, \cite{metzer2023latent-lucid-29}, \cite{michel2022text2mesh-lucid-31} aims to generate 3D content without 3D training data, enabling the conversion from concept to reality. However, zero-shot text-to-3D generation tasks~\cite{chen2019text2shape-2405-1}, \cite{duan2023dynamic-2405-4}, \cite{duan2020sofa-2405-5}, \cite{wu2016learning-2405-40} are constrained by the inherent complexity of the wild world and the scarcity of 3D data, unlike text-to-image~(T2I) tasks~\cite{ruiz2023dreambooth-2405-30}, \cite{saharia2022photorealistic-2405-32}. From this perspective, generating high-quality 3D content from text is still a significant challenge. 

\begin{figure}
    \centering
    \includegraphics[width=1\linewidth]{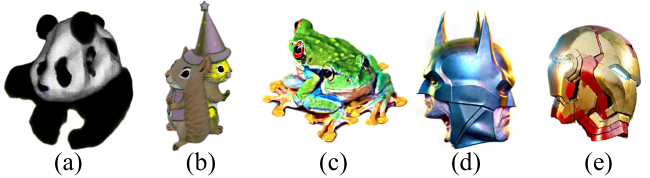}
    \caption{Examples of the Multi-Face Janus Problem: (a) generated by DreamFusion~\cite{poole2022dreamfusion-lucid-34}, (b) by SDI~\cite{lukoianov2024score}, (c) by LucidDreamer~\cite{liang2024luciddreamer-lucid}, (d) by DreamScene~\cite{10.1007/978-3-031-72904-1_13-dreamscene}, and (e) by GE3D~\cite{li2024text}.}
    \label{Fig_1}
\end{figure}

Earlier works~\cite{mildenhall2021nerf-lucid-32}, \cite{shi2023mvdream-lucid-41}, \cite{poole2022dreamfusion-lucid-34}, \cite{hong2023debiasing-disbais}, \cite{miao2025laser} based on Neural Radiance Fields (NeRF) introduce a framework that extended the 2D generative capabilities of diffusion models into the 3D domain by leveraging a view-conditioned projection mechanism. For instance, DreamFusion~\cite{poole2022dreamfusion-lucid-34} introduces the Score Distillation Sampling (SDS) technique, effectively aligning the rendering of 3D content with 2D generative priors. However, rendering in NeRF primarily relies on implicit 3D representations, which significantly increases computational cost and makes it challenging to produce high-resolution images~\cite{liang2024luciddreamer-lucid}. Recently, 3D Gaussian Splatting~\cite{kerbl20233d-2405-13} has been introduced as an efficient alternative to NeRF’s implicit 3D representation, enabling real-time, high-resolution rendering through explicit point-based modelling. Its adoption in State-of-the-Art~(SOTA) zero-shot text-to-3D generation methods~\cite{liang2024luciddreamer-lucid}, \cite{sun2024gseditpro-3dgswork1}, \cite{di2024hyper-hyper3d}, \cite{10.1007/978-3-031-72775-7_16-gcs} highlights its effectiveness in delivering superior results in quality and speed. However, these recent methods still struggle with generation precision, e.g., the multi-face Janus problem~\cite{shi2023mvdream-lucid-41}, \cite{poole2022dreamfusion-lucid-34}. A single 3D object exhibiting inconsistencies across its faces in different views, as shown in Fig.~\ref{Fig_1}, significantly undermines the realism and coherence of the 3D output. Therefore, ensuring view consistency in 3D generation is crucial for high-quality results, which is the primary motivation for this paper.

\begin{figure*}
    \centering
    \includegraphics[width=0.99\linewidth]{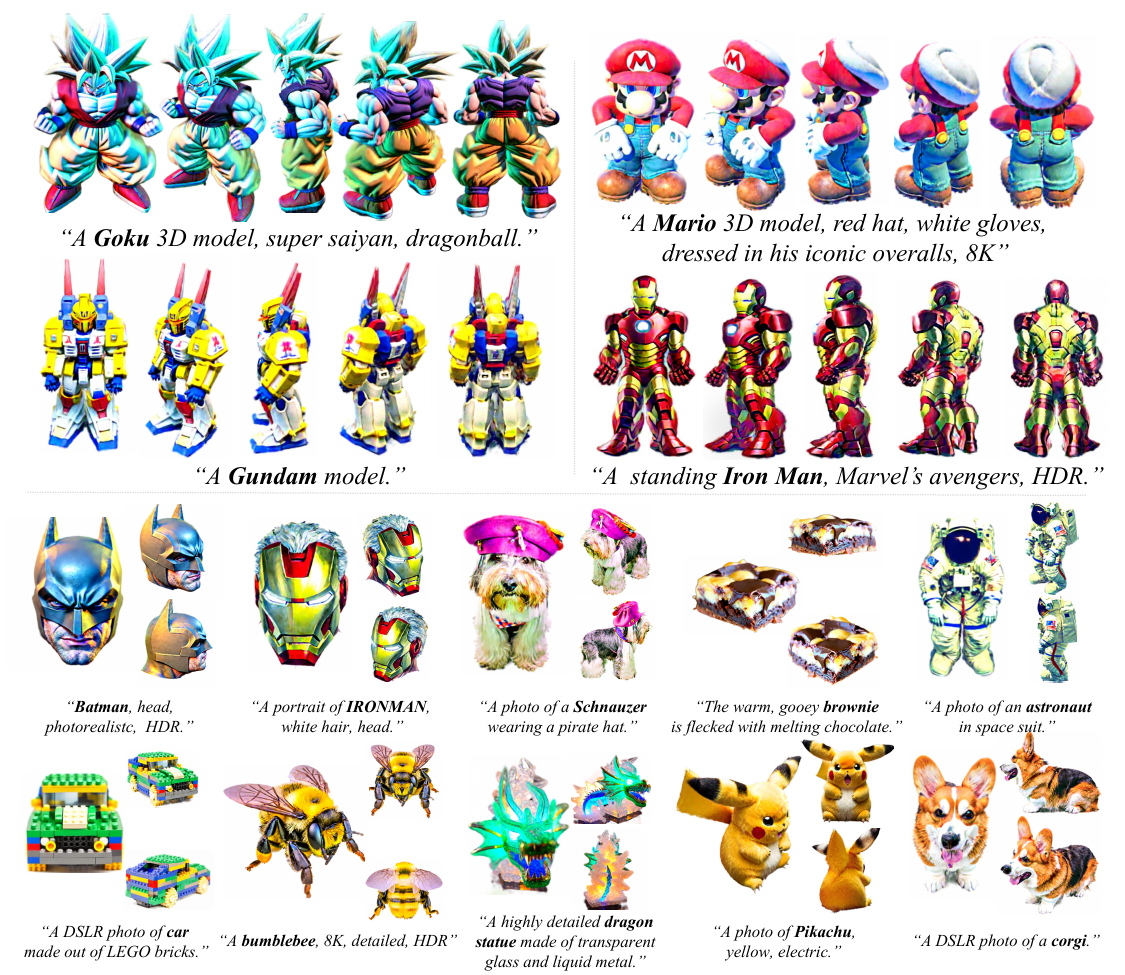}
    \caption{Examples of text-to-3D content creation using our method. We introduce $ConsDreamer$, a method that guarantees multi-view consistency by leveraging a View Disentanglement Model and a novel partial order loss to ensure semantic clarity across views (detailed in Section~\ref{sec:Methodology}). The generative 3D results demonstrate the superiority of ConsDreamer. Please zoom in for details.}
    \label{Fig_2}
\end{figure*}

Considering the most recent text-to-3D methods, they predominantly rely on T2I models to generate reference images. However, these models are typically trained on datasets sourced from publicly available resources and online platforms, such as photo-sharing sites. A notable limitation of these datasets is the lack of diverse and well-annotated multi-view images, which results in a bias toward high-frequency single view text-image pairs. As a consequence, T2I models tend to prioritise certain viewpoints, particularly frontal or side views~\cite{armandpour2023re-perpneg}. This view bias adversely impacts the consistency of 3D generation. When given rendered images and corresponding textual descriptions specifying particular views, the model often defaults to its prior view preference, rather than adhering to the exact view that training required. This misalignment leads to multi-view inconsistencies, where features across different views are incompatible, exacerbating the multi-face Janus problem in text-to-3D generation. Furthermore, the dominance of preferred viewpoints introduces an unbalanced view distribution, where even in the absence of explicit view conditions, the model still favours certain perspectives.

Perp-Neg~\cite{armandpour2023re-perpneg} is currently the most widely adopted strategy to mitigate such view biases. While it alleviates partial view conflicts via orthogonal decomposition in the denoising score space, it has a fundamental limitation: it only operates on the final output of the UNet, and cannot eliminate the aggregated prior view features that are already formed and accumulated throughout the UNet’s internal denoising pipeline (See Section~\ref{sec-Analysis of the multi-face Janus problem.} for detailed analysis). As a result, the multi-face Janus artefacts still persist even with Perp-Neg enabled. To tackle the aforementioned problem, we first conduct a mathematical analysis of the Janus issue. This analysis reveals that the prior view biases in T2I models affect both the conditional and unconditional terms used in existing text-to-3D methods. The unconditional term, which is independent of the textual prompt, is directly shaped by these inherent view biases. In the conditional term, prompt embeddings are fused with the target view description and then fed into the UNet as conditioning signals to generate the specified view. However, when the prior view preference of the T2I model disagrees with the desired target view, their interaction leads to ambiguous view semantics and unstable gradients. Therefore, we propose a View Disentanglement Module (VDM) that removes prior view bias and explicitly injects target-view control before the denoising process, thereby avoiding the accumulation of prior-view features inside the UNet. VDM computes subject-keyword view-specific residuals from the original prompt and the view-augmented prompt, and the residuals are then used to cancel prior view bias and to inject the desired target-view information according to the camera azimuth. In this way, the conditioning representation entering the UNet already aligns with the user-specified view, leading to a substantial reduction in Janus artefacts in multi-view text-to-3D generation.

For the unconditional term, we introduce a similarity-based partial order loss $\mathcal{L}_\text{P}$, which leverages the inherent relationship between view similarity and azimuthal angle distances. To implement this, a Cartesian coordinate framework is established, and an expected similarity partial order is determined based on the proximity of the azimuthal angles of multiple views to a reference view. The actual cosine similarity scores of the rendered-image features are constrained to conform to this partial order. By enforcing these constraints, $\mathcal{L}_\text{P}$ ensures cross-view consistency in 3D content, effectively addressing the multi-face Janus problem. We also conducted comprehensive experiments to illustrate the efficacy of our method, ConsDreamer, which confirms that it can better alleviate the multi-face Janus problem, strengthen inter-view semantic correlations, and enhance the quality of generated content. The main contributions of this work are summarized as follows:

\begin{itemize}
\item[$\bullet$] We begin by examining the fundamental principles of T2I models to formulate a mathematical framework for text-to-3D generation. This framework provides an in-depth analysis of the root causes of the multi-face Janus problem. Our analysis identifies two key problematic terms—conditional and unconditional—that introduce prior view biases during the score distillation process. To address these issues, we propose optimization strategies for both terms.
\item[$\bullet$] We propose a View Disentanglement Model~(VDM) to eliminate prior view preferences and integrate target view control in the conditional term. For the unconditional term, we introduce a similarity-based partial order loss to enhance the model’s view-awareness. Together, these components synergistically improve the clarity of view semantics.
\item[$\bullet$] We conduct extensive experiments on a wide range of prompts to verify the effectiveness of the proposed method when integrated into different 3D representations and various score-distillation strategies.
\end{itemize}

\section{Related Works}

\subsection{\textit{Differentiable 3D Representations}}

Differentiable 3D representation methods are especially essential for zero-shot text-guided 3D generation. Their differentiable nature enables a trainable pathway, optimizing 3D parameters by ensuring that rendered results from random views align with the prompt inputs. Previously, various representations are introduced for text-to-3D generation~\cite{mildenhall2021nerf-lucid-32}, \cite{barron2021mip-lucid-3}, \cite{ge2023ref-lucid-8}, \cite{shen2021deep-lucid-40}. Among these, NeRF~\cite{mildenhall2021nerf-lucid-32} emerges as the most widely adopted representation in text-to-3D generation tasks. NeRF enables novel view synthesis from unobserved viewpoints and leverages guidance information from the T2I model to optimize the tri-plane~\cite {shue20233d-triplane} parameters through a pre-trained MLP network, demonstrating broad applicability across downstream 3D tasks~\cite{lin2023magic3d-lucid-21}, \cite{mildenhall2021nerf-lucid-32}, \cite{shi2023mvdream-lucid-41}, \cite{poole2022dreamfusion-lucid-34}, \cite{hong2023debiasing-disbais}. Despite its success, NeRF faces challenges in rendering speed and memory consumption due to its implicit nature. A new 3D representation method, named 3D Gaussian Splatting, is thus introduced. Gaussian Splatting-based methods~\cite{kerbl20233d-2405-13}, \cite{liang2024luciddreamer-lucid}, \cite{tang2023dreamgaussian-lucid-45}, \cite{liu2024sherpa3d-sherpa3d} address these limitations using explicit 3D representations with anisotropic Gaussians. They ensure high reconstruction quality, real-time rendering, and reduced computational demands, making them ideal for large-scale and dynamic scenes. By replacing the NeRF method with the explicit 3D Gaussian Splatting representation, it becomes possible to integrate pre-trained 3D point cloud models like Point-E~\cite{nichol2022point-lucid-33} and \mbox{Shape-E}~\cite{jun2023shap-sherpa3d-28} into the 3D generation framework, providing a prior 3D knowledge foundation.

\subsection{\textit{Text-to-3D Generation}}

Early efforts, such as DreamField~\cite{jain2022zero-2405-10}, leverage the cross-modal information of CLIP~\cite{ramesh2021zero-2405-26} to translate text into 3D. With the advent of diffusion models, a new paradigm~\cite{poole2022dreamfusion-lucid-34}, \cite{tang2023dreamgaussian-lucid-45} emerges for extracting 3D assets from pre-trained T2I models, leveraging 2D diffusion model outputs to guide differentiable 3D representations. For example, DreamFusion introduces Score Distillation Sampling (SDS), which aligns 3D rendering information with diffusion priors and optimises model parameters through backpropagation. However, SDS-based approaches often introduce smoothing effects. Recent research~\cite{wu2016learning-2405-40}, \cite{katzir2023noise-lucid-18}, \cite{wang2024prolificdreamer-lucid-47}, \cite{yu2023text-lucid-49}, \cite{zhu2023hifa-lucid-52}, \cite{10.1007/978-3-031-72667-5_1-scaledreamer}, \cite{liang2024luciddreamer-lucid} aims to address this. LucidDreamer~\cite{liang2024luciddreamer-lucid} addresses over-smoothing issues via Interval Score Matching (ISM), integrating Denoising Diffusion Implicit Models~(DDIM)~\cite{saharia2022photorealistic-2405-32} inversion to enable deterministic sampling. Subsequent advancements, DreamerXL~\cite{miao2024dreamer-dreamxl}, ExactDreamer~\cite{zhang2024exactdreamer-exactdreamer}, and Guided Consistency Sampling (GCS)~\cite{10.1007/978-3-031-72775-7_16-gcs} further refine this paradigm by systematically reducing accumulated errors in DDIM’s reverse process. 

Recent advances also explore feed-forward methodologies~\cite{nichol2022point-lucid-33, jun2023shap-sherpa3d-28, xu2023dmv3d, liu2024pi3d, cao2023large-difftf, xiang2024structured}, which learn end-to-end text-to-3D mappings via paired data training to enable one-shot fast 3D generation with high in-distribution consistency. These methods, however, demand substantial GPU resources for additional training and require extensive 3D datasets. Furthermore, the overly conventional object categories in existing 3D training datasets severely limit the open-vocabulary generalisation capability of these feed-forward methodologies~\cite{jiang2024surveytextto3dcontentsgeneration}, and their inherent end-to-end black-box generation paradigm also leads to weak controllability and great difficulty in fine-grained editing.

\subsection{\textit{Mitigating the Multi-Face Janus Issue}}

Several studies have specifically tackled the multi-face Janus problem. Among these, Perp-Neg~\cite{armandpour2023re-perpneg} stands out as the most widely adopted, which leverages the geometric properties of the score space to enhance the utilisation of negative prompts. However, this method operates solely on the UNet output, thereby neglecting the highly aggregated prior view information throughout the UNet processing pipeline. Consequently, the multi-face Janus problem may still recur when lifting 2D representations to 3D. MVDream~\cite{shi2023mvdream-lucid-41}, Zero123~\cite{liu2023zero1to3}, and SV3D~\cite{voleti2024sv3d} all rely on pre-trained models and finetuning, which introduces high computational costs and dependency on large datasets. While MVDream ensures view consistency through fine-tuning, and Zero123 and SV3D generate multi-view 3D representations, these methods often struggle with maintaining consistent 3D geometry and texture across views. Moreover, they require extensive training, making them less efficient for quick deployment. In this paper, we introduce a novel method that eliminates prior view biases by disentangling view features and ensuring multi-view consistency through a carefully designed partial order loss. This approach significantly enhances view semantic clarity without the need for additional training or 3D datasets, offering a versatile, rapid-deployment solution that can be easily applied to a wide range of 3D generation frameworks.

\begin{figure*}[h]
    \centering
    \includegraphics[width=0.9 \linewidth]{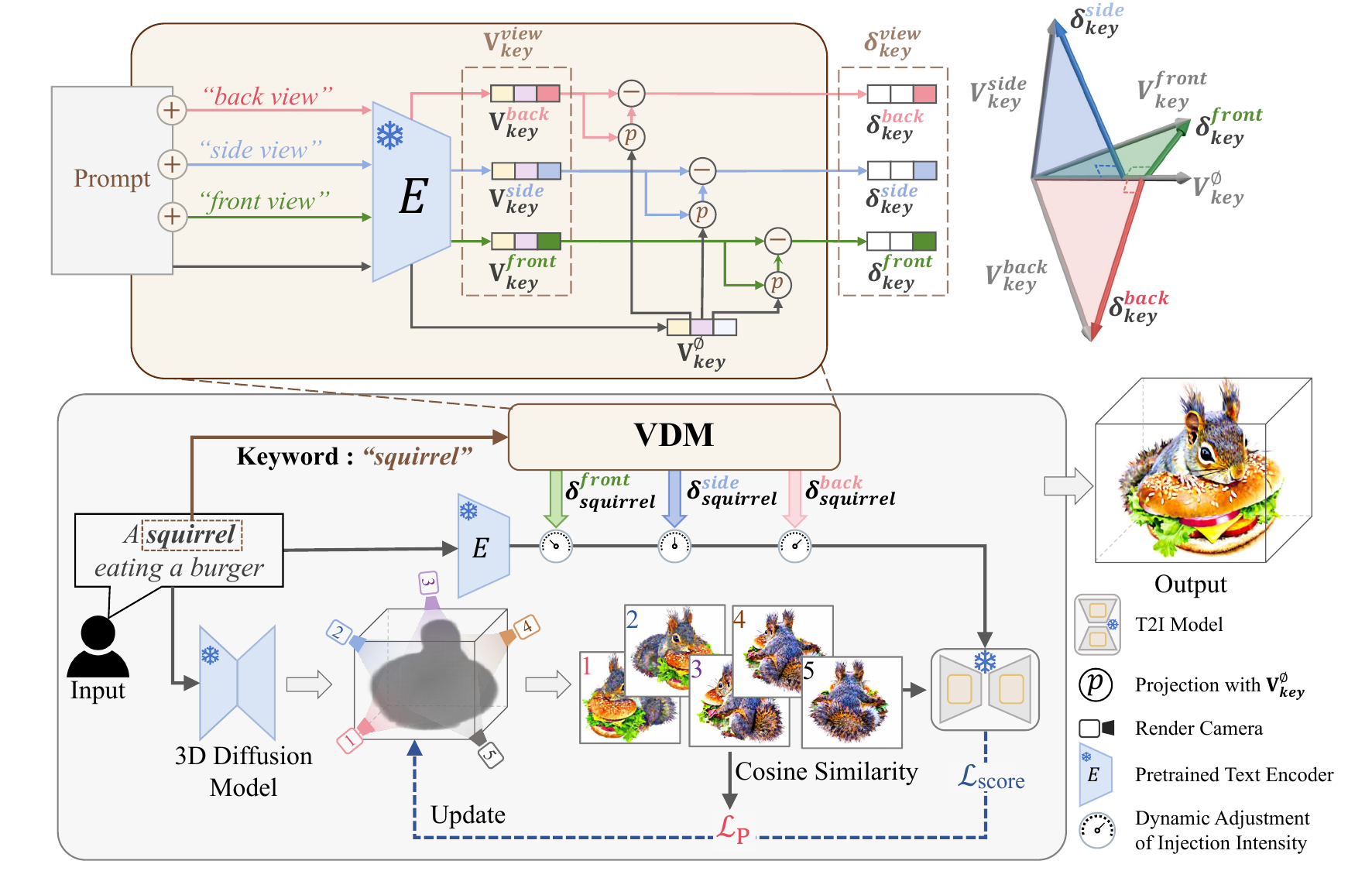}
    \caption{ An overview of ConsDreamer. Our method is built upon the main flow of 3D content distilled from the T2I model. ConsDreamer introduces two key innovations: (a) VDM disentangles the keyword in the prompt to obtain the canonical view features \( \delta_{\text{key}}^{\text{view}} \), which are then used for precise view control (detailed in Section~\ref{sec-View Disentanglement Module.}). (b) A novel partial order loss $\mathcal{L}_\text{P}$ is introduced among multi-view rendered images to endow the model with view-aware capabilities (detailed in Section~\ref{Partial Order Loss for Cross-View Consistency}). Together, the VDM and $\mathcal{L}_\text{P}$ enhance the clarity of view semantics and significantly mitigate the multi-face Janus problem.}
    \label{Fig_3}
\end{figure*}

\section{Methodology}
\label{sec:Methodology}

\begin{table}[t]
\centering
\footnotesize  
\caption{Summary of main symbols and notations used in the analysis.}
\setlength{\tabcolsep}{3pt} 
\begin{tabular}{p{0.18\linewidth} p{0.74\linewidth}} 
\toprule
Symbol & Description \\
\midrule
$c$ & User-provided text prompt. \\
$c_{sbj}$ & Subject-related component of $c$. \\
$c_v$ & Prior view bias component of $c$. \\
$\lambda$ & View condition in diffusion model. \\
$\lambda_n$ & View condition of $n$-th optimization step. \\
$r$ & Azimuth angle of camera for $\lambda_n$. \\
$\Omega$ & Set of views with strong priors, $\Omega=\{\text{front},\text{side}\}$. \\
$R_n$ & 2D image rendered from 3D representation at step $n\!-\!1$. \\
$\mathcal{L}_{\text{P}}$ & Similarity-based partial order loss. \\
\midrule
$\boldsymbol{z}_t$ & 2D latent at diffusion step $t$, $t\in\{0,\ldots,T\}$. \\
$\boldsymbol{z}_0$ & Clean 2D latent (reverse diffusion output). \\
$\boldsymbol{z}_T$ & Final noise latent sampled from $p(\boldsymbol{z}_T)$. \\
$T$ & Total diffusion time steps. \\
$p(\cdot)$ & Probability density function. \\
$p_{2D}(\boldsymbol{z}_0\mid c)$ & 2D latent distribution (conditioned on $c$). \\
$\tilde{p}_{3D}(\theta)$ & Unnormalized PDF of 3D representation. \\
$N$ & Number of optimisation iterations. \\
$\theta$ & 3D parameters. \\
$Z_{\theta,n}$ & 3D representation at iteration $n$ (parameterized by $\theta$). \\
\midrule
$\nabla_{\theta}$ & Gradient operator w.r.t. $\theta$. \\
$M$ & Compatibility term ($c_v$ vs. $\lambda_n$, Eq.~\eqref{eq11}). \\
$\langle\cdot,\cdot\rangle$ & Inner product of two vectors. \\
$\|\cdot\|$ & Euclidean norm of a vector. \\
$|\cdot|$ & Absolute value of a scalar. \\
$\perp$ & Orthogonality (e.g., $c_{sbj}\perp c_v$). \\
\midrule
$\mathbf{V}^{\emptyset}_{\text{key}}$ & Keyword embedding (no view description). \\
$\mathbf{V}^{\text{view}}_{\text{key}}$ & Keyword embedding (with specific view description). \\
$\delta^{\text{view}}_{\text{key}}$ & View-specific residual feature. \\
$Proj_v(u)$ & Projection of $u$ onto $v$: $Proj_v(u)=\frac{u\cdot v}{\|v\|^2}v$. \\
\bottomrule
\end{tabular}
\label{Tab_1}
\end{table}

To tackle the multi-face Janus problem, we propose ConsDreamer, a unified method to enhance view semantic clarity. Fig.~\ref{Fig_3} provides an overview of our approach. Section~\ref{sec-Analysis of the multi-face Janus problem.} presents a mathematical analysis of the problem, identifying two problematic terms: the conditional and unconditional terms, both of which introduce prior view preferences during the score distillation process. Based on this analysis, Section~\ref{sec-View Disentanglement Module.} introduces VDM to eliminate prior view preferences and integrate target view control in the conditional term. Meanwhile, for the unconditional term, Section~\ref{Partial Order Loss for Cross-View Consistency} explores similarity distribution patterns across views in rendered images and proposes a similarity-based partial order loss $\mathcal{L}_\text{P}$ to enhance spatial consistency in 3D content. For clarity, the main symbols and notations used in this section are summarised in Table~\ref{Tab_1}.

\subsection{\textit{Analysis of the multi-face Janus problem}}
\label{sec-Analysis of the multi-face Janus problem.}

In the Stable Diffusion framework, image generation is formulated as a reverse process that reconstructs the initial state $z_{0}$ from the final noise $z_{T}$. Guided by user prompts $c$, this reverse recovery models the probability distribution $p_{2D}(z_0 \mid c)$ for generating the final 2D latent representation $z_{0}$, as expressed in the following equation:
\begin{equation}		
\begin{aligned}
p_{2 D}\left(\boldsymbol{z}_{0}\!\mid\!c\right)\!=\!\int\! p\left(\boldsymbol{z}_{T}\right)\!\prod_{t=1}^{T}\! p\left(\boldsymbol{z}_{t\!-\!1} \!\mid\! \boldsymbol{z}_{t}, c\right) d{\boldsymbol{z}_{T}}\!\ldots\! d{\boldsymbol{z}_{1}},
\end{aligned}
 \label{eq3}
\end{equation}
where $p\left(\boldsymbol{z}_{t-1} \mid \boldsymbol{z}_{t}, c\right)$ denotes the probability of the prior latent representation $\boldsymbol{z}_{t-1}$ based on the current latent representation $\boldsymbol{z}_{t}$ and the user prompt $c$.

Furthermore, by injecting the views condition $\lambda$ into Eq.~\eqref{eq3}, the latent representations of 2D images with different views are obtained iteratively, as Eq.~\eqref{eq4-1} shows:
\begin{equation}		
\begin{aligned}
p_{2 D}\left(\boldsymbol{z}_{0} \!\mid\! c, \lambda\right)
\!=\!\int \!p\left(\boldsymbol{z}_{T}\right) \!\prod_{t=1}^{T}\! p\left(\boldsymbol{z}_{t\!-\!1} \!\mid\! \boldsymbol{z}_{t}, c, \lambda\right) d{\boldsymbol{z}_{T}} \!\ldots\! d{\boldsymbol{z}_{1}}.
		\end{aligned}
        \label{eq4-1}
\end{equation}
The corresponding 3D representation is the joint probability distribution of 2D representations of $N$ iterations. In other words, the multiplication of $n$ latent representation probabilities constructs an unnormalized probability density function for a 3D Gaussian model with parameters $\theta$:
\begin{equation}		
\resizebox{0.9\hsize}{!}{$\begin{aligned}
\tilde{p}_{3 D}(\theta)&\!=\!\prod_{n=1}^N \!\int\! p\left(\boldsymbol{z}_T\right) \!\prod_{t=1}^T\! p\left(\boldsymbol{z}_{t-1} \!\mid\! \boldsymbol{z}_t, c, \lambda_n\right) d_{\boldsymbol{z}_T} \!\ldots\! d_{\boldsymbol{z}_1}\\
&=\!\prod_{n=1}^N\! p_{2 D}\left(\boldsymbol{z}_{0} \!\mid\! c, \lambda_n\right),
		\end{aligned}$}
        \label{eq4}
\end{equation}
where $\lambda_n$ is the view of the $n^\text{th}$ iteration. It is noticeable that $p_{2D}\left(z_{0} \mid c, \lambda_n\right)$ in Eq.~\eqref{eq4} illustrates a reverse process given a view $\lambda$ in the diffusion model, and the diffusion model is fixed in training. For emphasis on the procedure from 2D distribution to 3D distribution, Eq.~\eqref{eq4} can be simplified as:
\begin{equation}		
\begin{aligned}
  \tilde{p}_{3 D}(\theta)=\prod_{n=1}^{N} p\left(Z_{\theta, n} \mid \lambda_{n}, c, R_n\right),
		\end{aligned}
        \label{eq5}
\end{equation}
where $R_n$ is a 2D rendered image from the 3D content in the ${(n-1)}^\text{th}$ iteration, and $Z_{\theta, n}$ denotes the 3D representation at the $n^\text{th}$ iteration. In Eq.~\eqref{eq5}, the user prompt $c$ is expected only to provide subject information without views information. However, the current pre-trained T2I models incorporate view prior knowledge when ``understanding” user prompt $c$ even if $c$ lacks explicit view information. This prior knowledge arises from the prior preference views of training data. Therefore, we decompose $c$ into a subject component $c_{sbj}$ and a view prior component $c_v$, which are expected to be orthogonal:
\begin{equation}		
\resizebox{0.9\hsize}{!}{$\begin{aligned}
\tilde{p}_{3 D}(\theta) = \prod_{n=1}^N p\left(Z_{\theta, n} \mid \lambda_n, c_{sbj}, c_v, R_n\right), \quad \text{s.t.} \, c_{sbj} \perp c_v,
		\end{aligned}$}
        \label{eq6}
\end{equation}
where $c_v=c-\frac{\left\langle c, c_{sbj}\right\rangle}{\left|\left|c_{sbj}\right|\right|^2} c_{sbj}$, $\left\langle \cdot, \cdot \right\rangle$ represents the inner product, and $\left|\left| \cdot \right|\right|$ denotes the Euclidean norm. Further, applying the logarithm to each side of Eq.~\eqref{eq6} yields:
\begin{equation}		
\begin{aligned}
{\rm \log} \tilde{p}_{3 D}(\theta)=\sum_{n=1}^N {\rm \log} p\left(Z_{\theta, n} \mid \lambda_n, c_{sbj}, c_v, R_n\right),
		\end{aligned}
        \label{eq7}
\end{equation}
the gradient of $\theta$ denoted as $\nabla_{\theta} {\rm \log} \tilde{p}_{3 D}(\theta)$, can be expressed as:
\begin{equation}		
\resizebox{0.90\hsize}{!}{$\begin{aligned}
 \nabla_{\theta} {\rm \log} \tilde{p}_{3 D}(\theta)=\sum_{n=1}^{N} \nabla_{\theta} {\rm \log} p\left(Z_{\theta, n} \mid \lambda_{n}, c_{sbj}, c_{v}, R_n\right)
\\=\sum_{n=1}^{N}\left(\frac{\partial {\rm \log} p\left(Z_{\theta, n} \mid \lambda_{n}, c_{sbj}, c_{v}, R_n\right)}{\partial_{Z_{\theta, n}}}\right) \frac{\partial_{Z_{\theta, n}}}{\partial_{\theta}},
		\end{aligned}$}
        \label{eq8}
\end{equation}
as $c_{sbj}$ and $c_v$ are independent, Eq.~\eqref{eq8} can be further expanded using Bayes' theorem. Specifically, the joint distribution $p(Z_{\theta, n}, \lambda_n, c_{sbj}, c_v, R_n)$ satisfies the decomposition property, allowing it to be expressed as $p(Z_{\theta, n}) p(\lambda_n, c_{sbj}, c_v, R_n \mid Z_{\theta, n})$. This ensures that the transformation using Bayes' theorem is valid:
\begin{equation}		
\resizebox{0.87\hsize}{!}{$\begin{aligned}
\nabla_\theta {\rm \log} \tilde{p}_{3 D}(\theta)
&=\sum\limits_{n=1}^N\!\left(\!\frac{\partial {\rm \log} p\left(\lambda_n, c_{sbj}, c_v, R_n \!\mid\! Z_{\theta, n}\right)}{\partial_{Z_{\theta, n}}}\right. \\
&\quad \left.+ \frac{\partial {\rm \log} p\left(Z_{\theta, n}\right)}{\partial_{Z_{\theta, n}}}\!-\!\frac{\partial {\rm \log} p\left(\lambda_n, c_{sbj}, c_v, R_n\right)}{\partial_{Z_{\theta, n}}}\right) \frac{\partial_{Z_{\theta, n}}}{\partial_\theta},
\end{aligned}$}
\label{eq8_full}
\end{equation}
the term $\frac{\partial {\rm \log} p\left(\lambda_n, c_{sbj}, c_v, R_n\right)}{\partial_{Z_{\theta, n}}}$ is a constant term independent of $Z_{\theta,n}$, and its derivative is zero, thus can be removed:
\begin{equation}		
\resizebox{0.9\hsize}{!}{$\begin{aligned}
\nabla_\theta {\rm \log} \tilde{p}_{3 D}(\theta)
\!=\!\sum\limits_{n=1}^N\!\left(\!\frac{\partial {\rm \log} p\left(Z_{\theta, n}\right)}{\partial_{Z_{\theta, n}}}+\!\frac{\partial {\rm \log} p\left(\lambda_n, c_{sbj}, c_v, R_n \!\mid\! Z_{\theta, n}\right)}{\partial_{Z_{\theta, n}}}\!\right)\! \frac{\partial_{Z_{\theta, n}}}{\partial_\theta},
\end{aligned}$}
\label{eq9}
\end{equation}
where $\frac{\partial {\rm \log} p\left(Z_{\theta, n}\right)}{\partial_{Z_{\theta, n}}}$ is the unconditional score~\cite{ho2020denoising-2405-7}, \cite{song2020score-2405-35} modelled by T2I models, intuitively reflects the model's perception of 3D objects without external view guidance. However, as previously mentioned, such perception is prone to prior preference views, which leads to the multi-face Janus problem. To address this issue, Section~\ref{Partial Order Loss for Cross-View Consistency} explores similarity distribution patterns across rendered images and proposes a similarity-based partial order loss $\mathcal{L}_\text{P}$ to enhance the model's view awareness in the absence of conditional view information.

Additionally, the second term in Eq.~\eqref{eq9} can be further expanded as follows:
\begin{equation}		
\resizebox{0.85\hsize}{!}{$\begin{aligned}
  &\frac{\partial}{\partial_{Z_{\theta, n}}} {\rm \log} p\left(\lambda_n, c_{sbj}, c_v, R_n \mid Z_{\theta, n}\right)\\
&\!=\!\underbrace{\frac{\partial}{\partial_{Z_{\theta, n}}} {\rm \log} p\!\left(c_{sbj}, c_v, R_n \!\mid\! Z_{\theta, n}\right)}_{\text{\normalsize Prior View Bias}}
\!+\!\underbrace{\frac{\partial}{\partial_{Z_{\theta, n}}} {\rm \log} p\!\left(\lambda_n, c_{sbj}, R_n \!\mid\! Z_{\theta, n}\right)}_{\text{\normalsize Target View Control}}\!+\frac{\partial}{\partial_{Z_{\theta, n}}} {\rm \log} M ,
		\end{aligned}$}
        \label{eq10}
\end{equation}
where $\frac{\partial}{\partial_{Z_{\theta, n}}} {\rm \log} p\!\left(c_{sbj}, c_v, R_n \!\mid\! Z_{\theta, n}\right)$ is independent of the view condition $\lambda_n$ and denotes the prior view bias term, while $\frac{\partial}{\partial_{Z_{\theta, n}}} {\rm \log} p\!\left(\lambda_n, c_{sbj}, R_n \!\mid\! Z_{\theta, n}\right)$ encodes view information exclusively governed by the view condition $\lambda_n$, corresponding to the target view control term, and $M=~\frac{p\left(\lambda_n, c_{sbj}, c_v, R_n \mid Z_{\theta, n}\right)}{p\left(c_{sbj}, c_v, R_n \mid Z_{\theta, n}\right) p\left(\lambda_n, c_{sbj}, R_n \mid Z_{\theta, n}\right)}$.

$c_{sbj}$ and $R_n$ are constants in the currentoptimisationn step, then $M$ can be simplified by using the definition of conditional probability:
\begin{equation}		
		\begin{aligned}
  M&=\frac{p\left(\lambda_{n}, c_{v} \mid Z_{\theta, n}\right)}{p\left(c_{v} \mid Z_{\theta, n}\right) p\left(\lambda_{n} \mid Z_{\theta, n}\right)}\\&=
  \frac{p\left(c_{v} \mid Z_{\theta, n}\right) p\left(\lambda_{n} \mid c_v,  Z_{\theta, n}\right)}{p\left(c_{v} \mid Z_{\theta, n}\right) p\left(\lambda_{n} \mid Z_{\theta, n}\right)}
  \\&=\frac{p\left(\lambda_{n} \mid c_v,  Z_{\theta, n}\right)}{p\left(\lambda_{n} \mid Z_{\theta, n}\right)}.
		\end{aligned}
        \label{eq11}
\end{equation}

\begin{figure}
    \centering
    \includegraphics[width=1\linewidth]{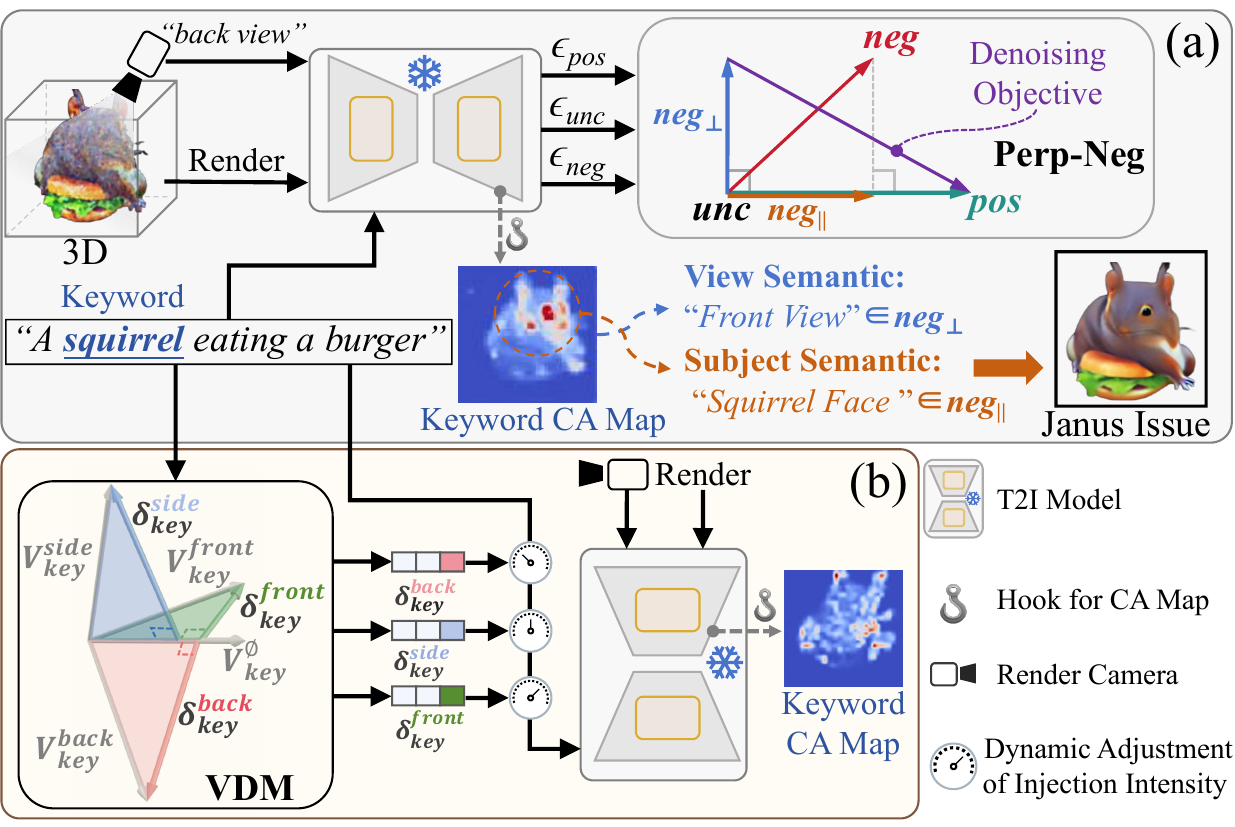}
    \caption{Comparison of PerpNeg and VDM in handling prior view biases. (a) Perp-Neg works \textbf{after} UNet denoising in the score space by orthogonalising negative-view scores. However, frontal-view features that have already been entangled into the subject-keyword CA map are absorbed into the preserved component $\textit{neg}_{\parallel}$, so prior-view faces survive and still cause Janus artefacts. (b) Our VDM instead intervenes \textbf{before} denoising in the prompt-embedding space: it extracts view-specific residuals from view-augmented prompts, subtracts prior-view residuals, and injects only the residual corresponding to the target azimuth, so the conditioning entering the UNet is already view-corrected, producing clean CA maps and substantially reducing Janus.}
    \label{Fig_4}
\end{figure}

Turning back to Perp-Neg, its score corresponding to the target view is treated as the positive condition $\textit{pos}$, while the score associated with an irrelevant view is treated as the negative condition $\textit{neg}$. The method computes an update direction from $\textit{neg}$ to $\textit{pos}$, and decomposes $\textit{neg}$ into a component parallel to $\textit{pos}$, $\textit{neg}_{\parallel}$, and an orthogonal component, $\textit{neg}_{\perp}$. The parallel part $\textit{neg}_{\parallel}$ is retained to preserve subject semantics, while $\textit{neg}_{\perp}$ is interpreted as view-specific content and is suppressed. While Perp-Neg alleviates partial view conflicts, its fundamental limitation can be rigorously interpreted and further revealed using our mathematical analysis framework above.

Observing Eq.~\eqref{eq10} and Eq.~\eqref{eq11}, when the view control $\lambda_n$ and the prior view preference $c_v$ conflict under the condition $Z_{\theta,n}$, i.e., $p\left(\lambda_n \mid c_v, Z_{\theta, n}\right) \ll p\left(\lambda_n \mid Z_{\theta, n}\right)$, the prior view bias and the target view control in Eq.~\eqref{eq10} will have a detrimental impact on the 3D generation jointly. In this scenario, even with Perp-Neg enabled, the cross-attention (CA) map of the subject-keyword (e.g., ``squirrel'') still exhibits strong frontal-view activations, as illustrated in Fig.~\ref{Fig_4} (a). These erroneous activations are incorrectly absorbed into $\textit{neg}_{\parallel}$ as subject semantics and thus preserved, which allows the multi-face Janus artefacts to persist. The distortions observed in the conditional results closely follow the spatial distribution of these abnormal facial activations. Additionally, in this case, the compatibility term $M$ approaches 0, and the third term $\nabla_{Z_{\theta,n}} {\rm \log} M$ in Eq.~\eqref{eq10} introduces a large negative gradient, further disrupting the optimisation process of the model. 

Therefore, establishing semantic consistency between $c_v$ and $\lambda_n$ from the source of prompt encoding is essential for effectively mitigating the multi-face Janus problem. Motivated by this, Section~\ref{sec-View Disentanglement Module.} proposes a View Disentanglement Module (VDM) to eliminate prior view preferences by disentangling view features and inject precise view control before the denoising process, thereby fundamentally enhancing the semantic consistency between $c_v$ and $\lambda_n$.

\subsection{\textit{View Disentanglement Module}}
\label{sec-View Disentanglement Module.}

Based on the mathematical analysis in Section~\ref{sec-Analysis of the multi-face Janus problem.}, we aim to eliminate the prior view preference and strengthen the target view specified by the camera parameters. The optimisation objective can be expressed as maximising the consistency between $c_{v}$ and $\lambda_{n}$:
\begin{equation}		
\begin{aligned}
\min _{\theta}\left|1-M\right|.
		\end{aligned}
        \label{eq12}
\end{equation}
As $M$ approaches 1, the model effectively understands the target view $\lambda_{n}$ specified by the rendering camera, independent of view biases in the pre-trained model. 

Optimizing Eq.~\eqref{eq12} is non-trivial, which motivates the proposal of the VDM to eliminate prior view biases and enhance the effectiveness of view control. A two-phase adjustment is applied to achieve the goal: the model’s prior view biases are removed from the original prompt $c$, and the view control is strengthened to ensure the generation process follows the user-specified target view.

 The process is illustrated in the upper of Fig.~\ref{Fig_3}. First, identify the keyword from the prompt (e.g., the keyword ``squirrel” of the prompt ``A squirrel eating a burger”), and extract the keyword embedding $\mathbf{V}_{{key}}^{\emptyset}$ from the prompt encoding results. Subsequently, new embeddings $\mathbf{V}_{{key}}^{{view}}$ corresponding to various views are obtained by combining the prompt with different view descriptions (e.g., ``back view,” ``side view,” and ``front view”). Unlike $\mathbf{V}_{{key}}^\emptyset$, $\mathbf{V}_{{key}}^{{view}}$ incorporates auxiliary contextual information, which enables the view-specific features $\delta_{{key}}^{{view}}$ isolating content-agnostic information. The process is formulated as follows:
\begin{equation}		
	\begin{aligned}
\delta_{ {key }}^{ {view }}=\mathbf{V}_{ {key }}^{ {view }}-{Proj}_{\mathbf{V}_{k e y}^\emptyset}\left(\mathbf{V}_{ {key }}^{ {view }}\right),
		\end{aligned}
        \label{eq13}
\end{equation}
where $Proj_v(u)=\frac{u \cdot v}{||v||^2} v$ represents the projection of $u$ onto $v$. The view information is deprived from $\mathbf{V}_{{key}}^{{view}}$ by projecting $\mathbf{V}_{{key}}^{{view}}$ to $\mathbf{V}_{{key}}^{\emptyset}$. Furthermore, the view features $\delta_{ {key }}^{ {view }}$ are obtained by subtracting this projection from the $\mathbf{V}_{{key}}^{{view}}$. The $\delta_{ {key }}^{ {view }}$ can be applied to user prompts, enabling view control.

\begin{figure}
    \centering
    \includegraphics[width=1\linewidth]{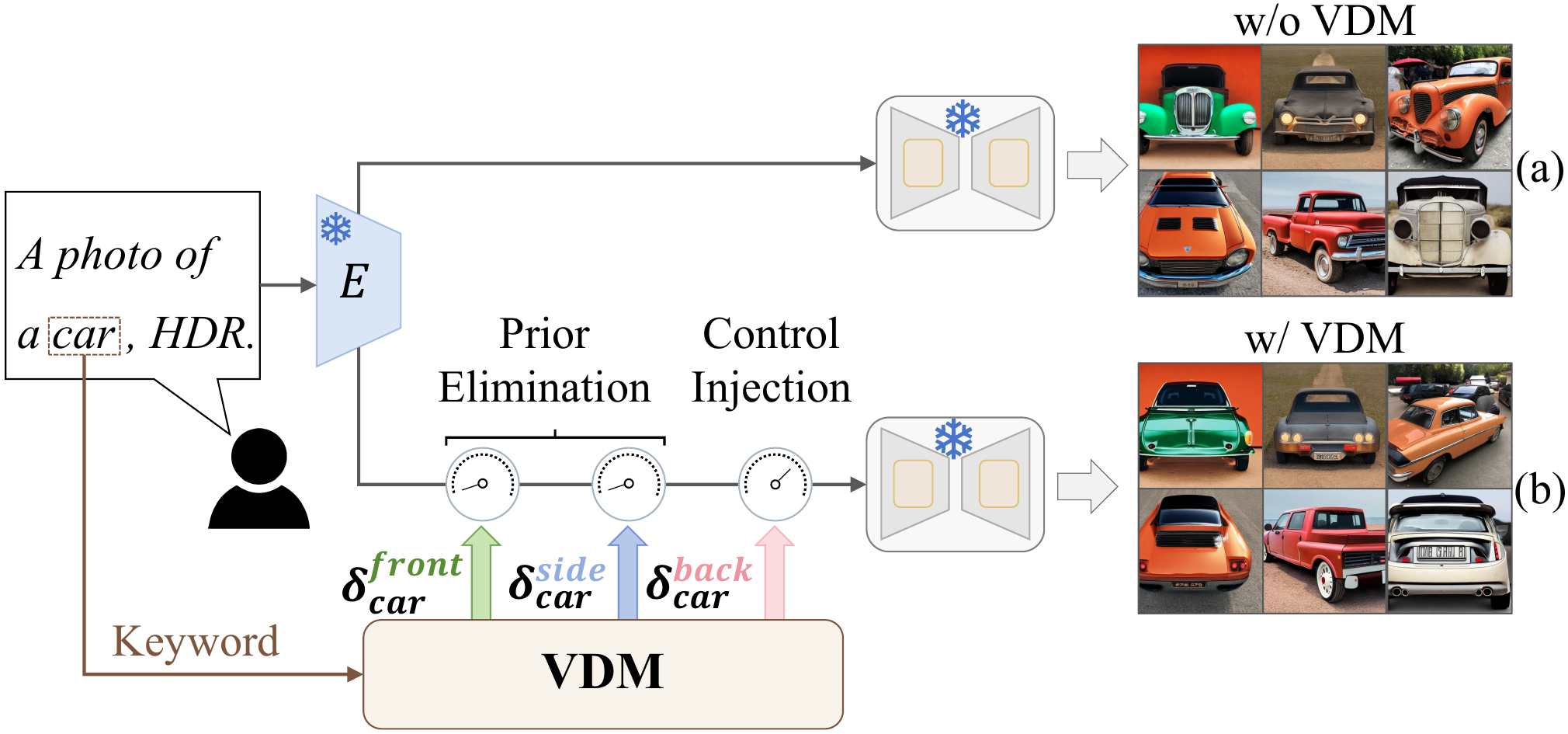}
    \caption{ Application of the VDM in 2D Generation. (a) Without the VDM, the model mainly generates front or side views of a car, while the back view is rarely generated. (b) With the VDM, prior view biases are eliminated and targeted view information is injected, allowing the model to successfully generate a car from the back view.}
    \label{Fig_5}
\end{figure}

In text-to-3D generation tasks, the prior view preferences within the set $\Omega=\{front, side\}$ need to be eliminated first. Subsequently, the view injection process is adaptively implemented using the azimuth $r$ at each optimization step and the intensity coefficient $w$. Specifically, when the azimuth $r$ falls within the range $(-90^\circ, 90^\circ)$, $\delta_{ {key }}^{ {side }}$ is injected to suppress prior view preferences from the front view:
\begin{equation}		
\begin{aligned}
\mathbf{V}^\emptyset_{\mathit{key}} \!\leftarrow\! 
\mathbf{V}^\emptyset_{\mathit{key}} \!- \!\sum^{\Omega}\! Proj_{\delta_{{key}}^{{view}}}(\mathbf{V}^\emptyset_{\mathit{key}})\! +\! w_1 \!\cdot \!\frac{|r|}{90} \!\cdot \!\delta^{\mathit{side}}_{\mathit{key}},
\end{aligned}
\label{eq3d1}
\end{equation}
where $w_1$ controls the intensity of view injection, $|r|/90$ normalizes the azimuth $r$ symmetrically within the range of -90° to 90°, ensuring that the injection intensity corresponds proportionally to the deviation from the front view. When the azimuth $r$ falls within the range $(-180^\circ, -90^\circ) \cup (90^\circ, 180^\circ)$, both $\delta_{{key}}^{{back}}$ and $\delta_{{key}}^{{side}}$ must be considered simultaneously, because diffusion models may be biased toward generating front views~\cite{armandpour2023re-perpneg}:
\begin{equation}		
\begin{aligned}
\mathbf{V}^\emptyset_{\mathit{key}} \!\leftarrow\! &\mathbf{V}^\emptyset_{\mathit{key}}\! - \!\sum^{\Omega}Proj_{\delta_{{key}}^{{view}}}(\mathbf{V}^\emptyset_{\mathit{key}}) \\&\!+\! w_2 \!\cdot\! \frac{|r| \!-\! 90}{90} \!\cdot \!\delta^{\mathit{back}}_{\mathit{key}}\! +\! w_3 \cdot \frac{180 \!- \!|r|}{90}\! \cdot \!\delta^{\mathit{side}}_{\mathit{key}},
\end{aligned}
\label{eq3d2}
\end{equation}
where $w_2, w_3$ control the intensity of view injection. As illustrated in Fig.~\ref{Fig_3}, we implement the dynamic adjustment of injection intensity using Eq.~\eqref{eq3d1} and Eq.~\eqref{eq3d2}, the model can more accurately ``understand" the target view $\lambda_n$ specified by the rendering camera. We fix $w_1=1.0$, $w_2=1.5$ and $w_3=1.0$ in all experiments, and results in Section~\ref{sec4:Experiments} demonstrate that this approach effectively mitigates the multi-face Janus problem.

\begin{figure}
\includegraphics[width=1\linewidth]{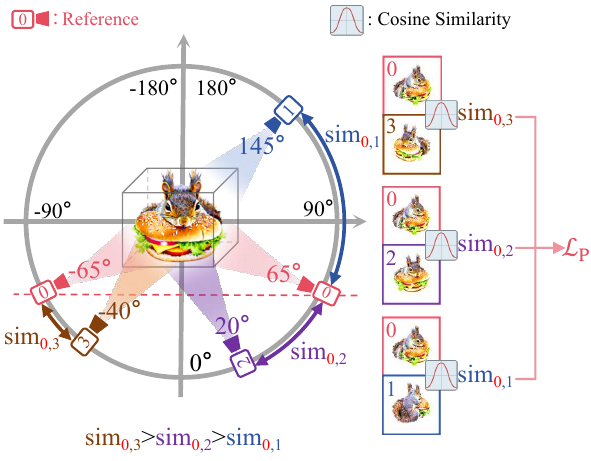}
    \caption{ Partial order loss for cross-view consistency. The random camera views are distributed on the unit circle based on their azimuth angles. A reference camera (e.g., camera 0) is selected, and a red dotted line parallel to the horizontal axis is constructed from this point to mirror the reference to a symmetric position. The azimuthal distances between other camera views and the nearest reference are computed to determine the expected similarity relationships ($\text{sim}_{0,3}>\text{sim}_{0,2}>\text{sim}_{0,1}$). The right-hand side of the figure computes the actual similarity relationships using cosine similarity, and $\mathcal{L}_\text{P}$ evaluates the alignment between the actual and expected similarity relationships.}
    \label{Fig_6}
\end{figure}

To further evaluate the effectiveness of the VDM, and given that prior view preferences in 3D tasks arise from similar biases in 2D generation, we also apply the VDM to 2D generation tasks. Fig.~\ref{Fig_5} illustrates the impact of VDM on 2D generation tasks. Typically, generated 2D results default to prior preference views. As shown in Fig.~\ref{Fig_5} (a), when the prompt is ``A photo of a car, HDR," the model predominantly generates front or side view images of a car, while images featuring a back view are rarely generated. To generate images from rare views like the ``back view," prior view preferences must be eliminated and target view information injected, as described in Eq.~\eqref{eq14}:
\begin{equation}
\scalebox{0.90}{$\displaystyle
\begin{aligned}
\mathbf{V}_{{key}}^{prompt} \!\leftarrow \!\mathbf{V}_{{key}}^{prompt} 
\underbrace{\!-\!\sum^{\Omega} Proj_{\delta_{{key}}^{{view}}}\!\left(\!\mathbf{V}_{{key}}^{prompt}\!\right)\!}_{\text{\scriptsize Prior Elimination}} 
\underbrace{+\eta \cdot \delta_{{key}}^{{back}}}_{\text{\scriptsize Control Injection}},
\end{aligned}
$}
\label{eq14}
\end{equation}
following Eq.~\eqref{eq13}, the VDM extracts the three view features: $\delta_{car}^{front}$, $\delta_{car}^{side}$, and $\delta_{car}^{back}$. The prior view bias is first eliminated using $\delta_{car}^{front}$ and $\delta_{car}^{side}$, and then the prompt embedding is enhanced by injecting $\delta_{car}^{back}$. As a result, the Stable Diffusion model successfully generates an image of a car from the back view, as shown in Fig.~\ref{Fig_5} (b). This demonstrates the effectiveness of the VDM in overcoming view biases and enabling precise view control in 2D generation tasks.

\begin{figure}
\includegraphics[width=1\linewidth]{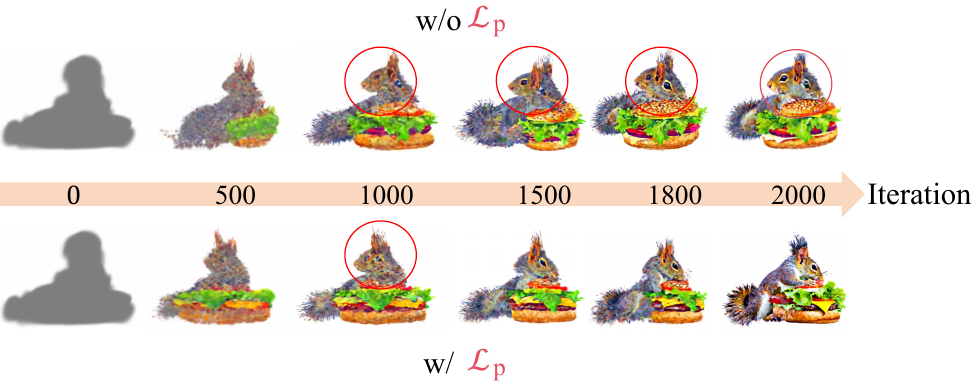}
    \caption{ Impact of $\mathcal{L}_\text{P}$ on 3D content over iterations. Using ``A squirrel eating a burger" as an example, $\mathcal{L}_\text{P}$ demonstrates a significant suppression effect on the multi-face Janus problem encountered during early training stages.}
    \label{Fig_7}
\end{figure}

\subsection{\textit{Partial Order Loss for Cross-View Consistency}}
\label{Partial Order Loss for Cross-View Consistency}

The VDM enhances the clarity of view semantics in user prompts, ensuring alignment between user prompts and generated objects in terms of view semantics. However, the mathematical analysis of the multi-face Janus problem in Eq.~\eqref{eq9} (Section~\ref{sec-Analysis of the multi-face Janus problem.}) reveals that the unconditional term, represented by the gradient $\frac{\partial {\rm \log} p\left(Z_{\theta, n}\right)}{\partial Z_{\theta, n}}$, lacks explicit guidance from view information. This absence of view-specific regulation makes the unconditional term more susceptible to introducing prior view biases into the generated content. During the iterative optimization process, $Z_{\theta, n}$ is projected as a 2D rendered image and input into the Stable Diffusion model, but this 2D projection does not preserve the specific view information in the 3D space. To address this, it is necessary to establish a connection between the unconditional guiding term and view control, ensuring that the unconditional term is regulated by the rendering camera parameters and infusing the model with view awareness.

\begin{figure*}
    \centering
    \includegraphics[width=1\linewidth]{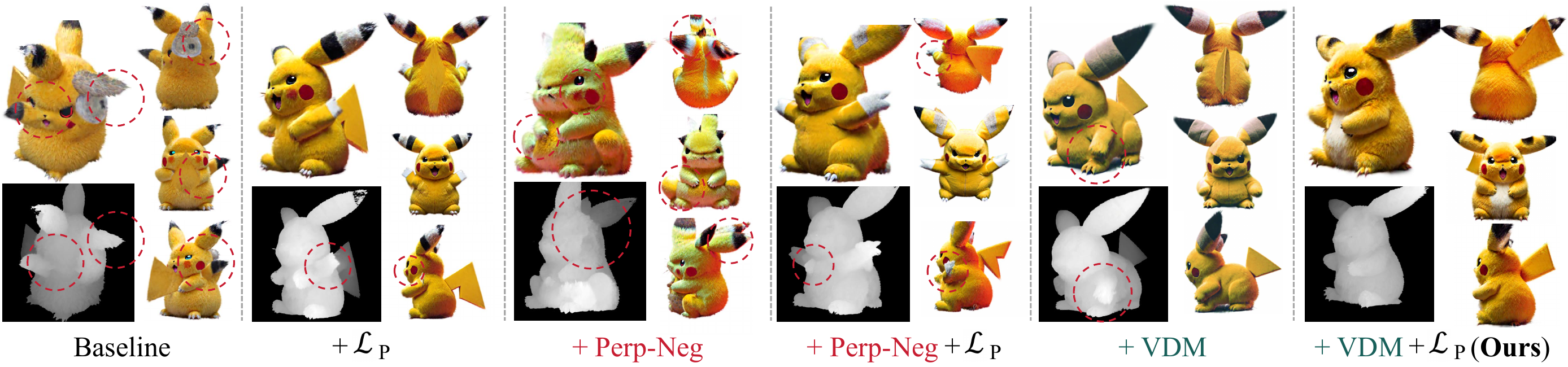}
    \caption{Qualitative ablation study results (prompt: ``A photo of a Pikachu, yellow, electric”). Red dashed circles mark inconsistent artefacts (e.g., redundant limbs, extra faces) in the generated results. Incrementally integrating our proposed method’s components into the baseline enhances generation consistency. To enhance the persuasiveness of our analysis, we include Perp-Neg as a variable.}
    \label{Fig_8}
\end{figure*}

To achieve this, we explore the relationship between the multi-view rendered images. First, each rendered image is encoded into high-dimensional semantic features, which capture the global view-aware semantics of the images and exhibit some robustness to flipping, rotation, and symmetry variations of the rendered content \cite{bengio2014representationlearningreviewnew}. A random image is selected from all rendered images as the reference image, and the cosine similarity between the other images and the reference image is computed at the feature level. As shown in Fig.~\ref{Fig_6}, the red camera represents the reference view selected from a series of randomly generated camera views $\Lambda$. It is evident that the closer the azimuth is to the reference azimuth, the higher the similarity score between the corresponding rendered image and the reference image. Furthermore, the scores gradually decrease symmetrically as the azimuth angle deviates from the reference. Therefore, we define a similarity-based partial order loss, where the specific ordering method is illustrated in Fig.~\ref{Fig_6}. A Cartesian coordinate system is established, and the random camera views are distributed on the unit circle based on their azimuth angles. One camera view is randomly chosen as the reference view, e.g., camera 0 in Fig.~\ref{Fig_6}, and a red dotted line parallel to the horizontal axis is constructed from this point. The reference camera is then mirrored to a symmetric point using this line. The azimuthal distances between other camera views and the nearest reference camera view are computed, determining the expected similarity partial order among the corresponding rendered images: $\text{sim}_{0,3}> \text{sim}_{0,2}> \text{sim}_{0,1}$. The similarity relationship between the features of rendered images obtained from Eq.~\eqref{eq9} must also conform to this order. Accordingly, the cosine similarity between the semantic features of corresponding rendered images is computed as the actual similarity and a similarity-based partial order loss $\mathcal{L}_\text{P}$ is employed to constrain the actual similarity to align with the expected similarity relationships. $\mathcal{L}_\text{P}$ is defined as:
\begin{equation}		
\begin{aligned}
\mathcal{L}_\text{P}\!=\!\sum_{i=1}^{|\Lambda|-2} \!\max \!\left(0, \text{sim}\!\left(F^{i+1}, F^0\right)\!-\!\text{sim}\!\left(F^i, F^0\right)\right),
\end{aligned}
\end{equation}
where $F^i$ is the encoded feature of the rendered image corresponding to the $i^{th}$ view in $\Lambda$, $F^0$ is the reference feature and $\operatorname{sim}\left(F^i, F^0\right)$ denotes the cosine similarity between $F^i$ and $F^0$. The $\mathcal{L}_\text{P}$ term can constrain the multi-face Janus problem early in training. As shown in Fig.~\ref{Fig_7}, using ``A squirrel eating a burger" as an example, without the $\mathcal{L}_\text{P}$ constraint, the generated result exhibits an irreconcilable multi-face Janus problem. When $\mathcal{L}_\text{P}$ is applied, the generated 3D content shows significant improvement starting from the 1000th iteration.

\begin{table*}[htbp]
\centering
\begin{threeparttable}
\caption{Quantitative Results of Ablation Study}
\label{Tab_2}
\begin{tabular}{l c c c c c c c c c c c c}
\toprule
\multirow{2}{*}{\textbf{Methods}} & \multicolumn{2}{c}{\multirow{2}{*}{\textbf{ImageReward}$ \uparrow$}} 
& \multicolumn{2}{c}{\multirow{2}{*}{\textbf{OpenCLIP} $\uparrow$}} & \multicolumn{4}{c}{\centering \textbf{A-LPIPS}} & \multicolumn{4}{c}{\centering \textbf{Frequency of Inconsistency}} \\
 \cmidrule(lr){6-9} \cmidrule(lr){10-13}
 & & & & & \multicolumn{2}{c}{\centering \textbf{VGG} $\downarrow$} & \multicolumn{2}{c}{\centering \textbf{Alex} $\downarrow$} & \multicolumn{2}{c}{\centering \textbf{f$_{\text{mf}}$(\%)} $\downarrow$} & \multicolumn{2}{c}{\centering \textbf{f$_{\text{inc}}$(\%)} $\downarrow$} \\
\midrule
 & Rank & Score & Rank & Score & Rank & Score & Rank & Score & Rank & Score & Rank & Score \\
\midrule
Janus Issue & 7 & -0.770 & \textbf{1} & \textbf{0.340} & 7 & 0.086 & 7 & 0.050 & 7 & 100.0 & 7 & 100.0 \\
Baseline & 6 & -0.637 & 2 & 0.338 & 6 & 0.079 & 6 & 0.047 & 6 & 85.7 & 6 & 97.1 \\
Baseline + $\mathcal{L}_\text{P}$ & 5 & 0.274 & 7 & 0.327 & 3 & 0.074 & 3 & 0.042 & 5 & 37.1 & 4 & 60.0 \\
Baseline + Perp-Neg & 4 & 0.285 & 6 & 0.328 & 4 & 0.076 & 5 & 0.043 & 4 & 24.0 & 5 & 64.0 \\
Baseline + Perp-Neg + $\mathcal{L}_\text{P}$ & 3 & 0.294 & 5 & 0.332 & 4 & 0.076 & 3 & 0.042 & 3 & 15.3 & 3 & 52.7 \\
Baseline + VDM & 2 & 0.465 & 3 & 0.333 & 2 & 0.068 & 2 & 0.039 & 2 & 11.4 & 2 & 42.0 \\
Baseline + VDM + $\mathcal{L}_\text{P}$ (Ours) & \textbf{1} & \textbf{0.707} & 3 & 0.333 & \textbf{1} & \textbf{0.060} & \textbf{1} & \textbf{0.035} & \textbf{1} & \textbf{3.4} & \textbf{1} & \textbf{31.0} \\
\bottomrule
\end{tabular}
\end{threeparttable}
\end{table*}

Our plug-and-play method integrates into multiple baseline frameworks with diverse score distillation strategies, unifying their score distillation losses into a consistent formulation $\mathcal{L}_\text{score}$. The final loss function is then formalised as:
\begin{equation}		
		\begin{aligned}
 { Loss }= \mathcal{L}_\text{score}+\kappa \mathcal{L}_\text{P},
		\end{aligned}
\end{equation}
where $\kappa$ is the weight coefficient of $\mathcal{L}_\text{P}$.

\section{Experiments}
\label{sec4:Experiments}

In this section, we provide a systematic evaluation of ConsDreamer through a dual approach integrating qualitative visual analysis and quantitative benchmarking. First, we conduct a comprehensive ablation study in Section~\ref{Ablation Studies} to evaluate the effectiveness of each individual module. Furthermore, Perp-Neg is additionally considered as an ablation variable to ensure a fair and comprehensive comparison. Next, in Section~\ref{Comparison with SOTA Methods}, ConsDreamer is integrated into six SOTA baselines encompassing diverse 3D representations (NeRF or 3DGS) and various score distillation strategies to validate the versatility of our method. Finally, we explore the improvement brought by VDM to T2I tasks in Section~\ref{Application of the VDM}, further extending the evaluation. We strictly follow the original hyperparameter settings of all baseline models~\cite{poole2022dreamfusion-lucid-34}, \cite{lukoianov2024score}, \cite{liang2024luciddreamer-lucid}, \cite{10.1007/978-3-031-72904-1_13-dreamscene}, \cite{li2024text}, including learning rate, iterations, and optimiser. To ensure fair comparison and full reproducibility, the same random seed and initial point cloud are used for all comparative experiments. All experiments were conducted using the Stable Diffusion v2.1 base model for distillation.

\begin{figure}
\includegraphics[width=1\linewidth]{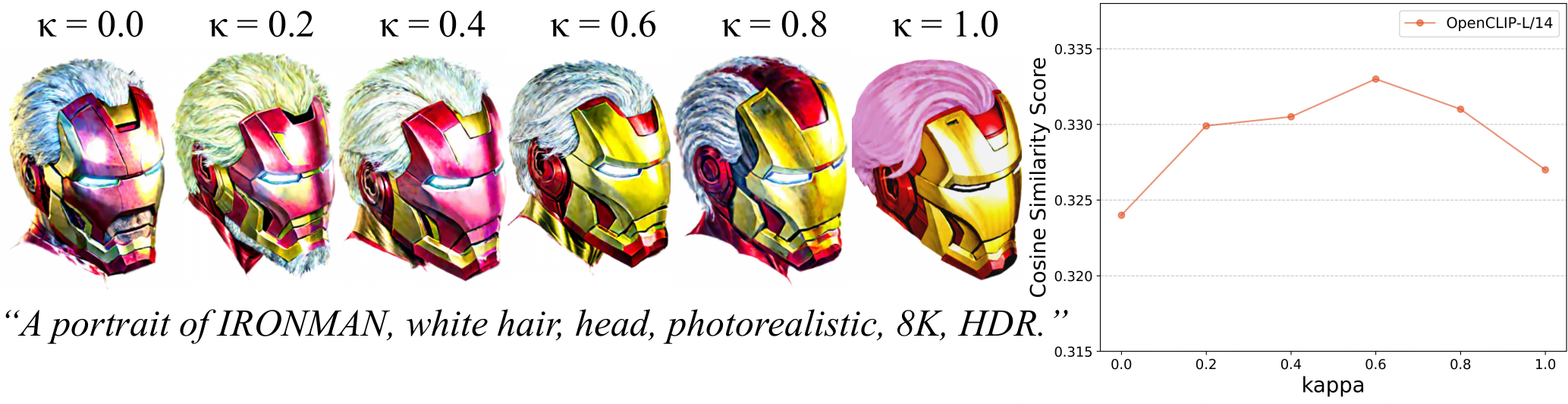}
    \caption{Ablation on weight coefficient $\kappa$ of $\mathcal{L}_\text{P}$. $\kappa$ = 0.6 yields the highest visual quality: excessively large values lead to over-smoothing and misalignment with textual prompts (e.g., ``white hair"), while overly small values cause view inconsistency in the generated results, resulting in poor fitting of the target object.}
\label{Fig_9}
\end{figure}

\subsection{\textit{Ablation Studies}}
\label{Ablation Studies}

\begin{figure}
    \centering
    \includegraphics[width=1\linewidth]{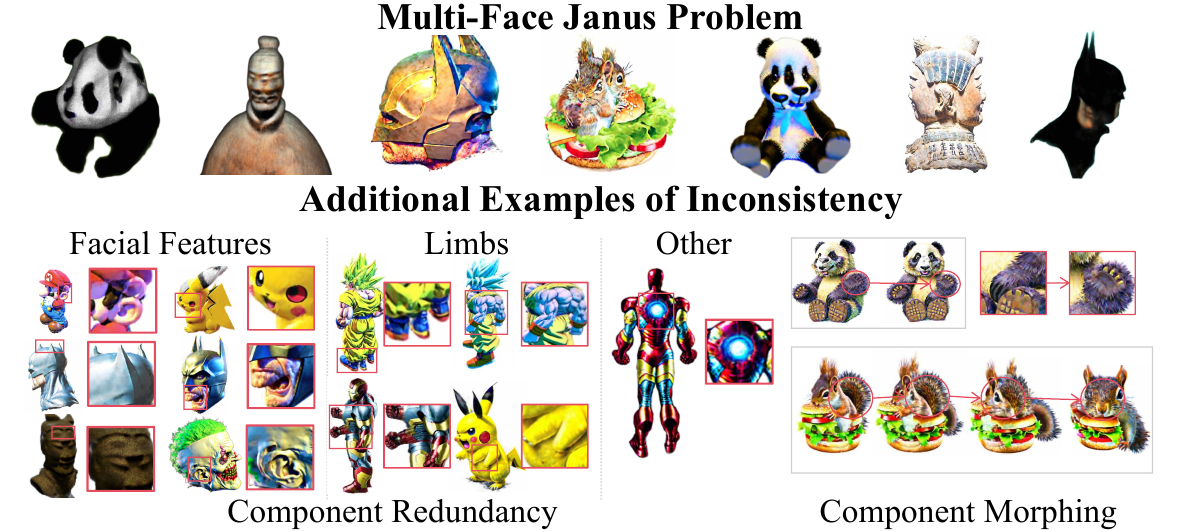}
    \caption{ Examples of inconsistencies in generated 3D content. The top section highlights the multi-face Janus issue. Additionally, the section below showcases other types of inconsistencies, including component redundancy and component morphing.}
    \label{Fig_10}
\end{figure}

\textit{1) Qualitative Results:}
By eliminating view semantic bias in T2I mappings and incorporating explicit consistency constraints across multi-view images, our method significantly improves the consistency of generated content. As illustrated in Fig.~\ref{Fig_8}, the baseline model without any consistency strategy suffers from severe multi-face Janus artefacts, which are characterised by additional limbs and redundant faces of Pikachu. When the perceptual loss $\mathcal{L}_\text{P}$ is applied, the extra limbs and ears are effectively suppressed. To investigate the performance gain of VDM over Perp-Neg, we also treat Perp-Neg as an ablation variable, and the visualisation results demonstrate a notable improvement with VDM achieving more consistent generation results than Perp-Neg. However, certain inconsistent artefacts, such as redundant limbs marked by red dashed circles, still persist. The model yields the optimal performance in terms of both consistency and visual quality when VDM is combined with $\mathcal{L}_\text{P}$. Furthermore, as shown in Fig.~\ref{Fig_9}, we conducted an ablation study on the weight coefficient $\kappa$ of the partial order loss $\mathcal{L}_\text{P}$.

\begin{figure}
    \centering
        \setlength{\abovecaptionskip}{0.cm}
	\setlength{\belowcaptionskip}{0.cm}
    \includegraphics[width=1\linewidth]{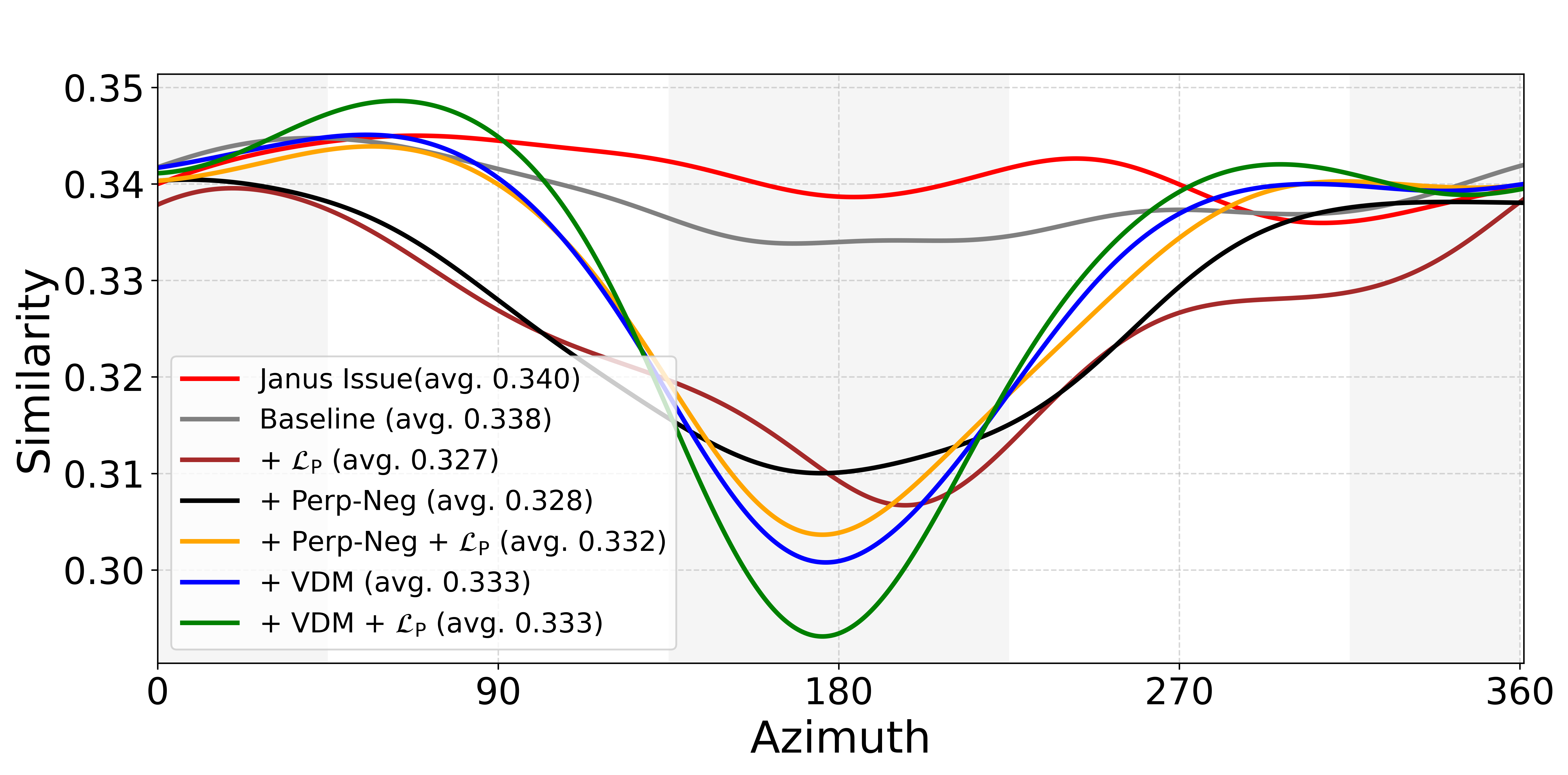}
    \caption{ Similarity distribution between prompts and rendered results across methods. The similarity distribution between prompts and rendered results for various methods is compared, the results demonstrate that our method generates 3D content with a similarity score distribution that aligns more closely with the ideal distribution.}
    \label{Fig_11}
\end{figure}

\textit{2) Quantitative Results:} Qualitative results visually demonstrate the effectiveness of the proposed method. We employed conventional evaluation metrics in text-to-3D generation, including ImageReward~\cite{NEURIPS2023_33646ef0-imagereward}, OpenCLIP-L/14~\cite{ramesh2022hierarchical-openclip}, and A-LPIPS~\cite{Zhang_2018_CVPR-alpips}, to objectively assess the effectiveness of our method. ImageReward measures the consistency between generated results and text descriptions, OpenCLIP-L/14 provides enhanced text-image alignment capabilities to capture fine-grained semantic information, and A-LPIPS evaluates the visual quality of generated images, focusing on perceptual similarity. Table~\ref{Tab_2} shows that each module of our method outperforms the scenario without intervention in various aspects, and the joint optimisation of both modules achieves even better performance. Additionally, we evaluated the percentage of generated results exhibiting \textbf{Inconsistency}, denoted as $\text{f}_\text{inc}$. We broadened the traditional definition of the multi-face Janus problem to encompass a wider range of inconsistencies, such as extra limbs and other implausible artefacts. Fig.~\ref{Fig_10} provides illustrative examples of \textbf{Inconsistency}. For a comprehensive evaluation, we selected a diverse set of prompts involving objects with distinct front-back differences (e.g., portraits, animals, and vehicles). The quantitative results are summarised in Table~\ref{Tab_2}, where the frequency of the multi-face Janus problem is denoted as $\text{f}_\text{mf}$. The results indicate that each proposed component significantly improves the consistency of the generated content. Notably, when directly comparing VDM against Perp-Neg, VDM demonstrates substantially superior performance, achieving an additional 52.5\% reduction in $\text{f}_\text{mf}$ (from 24.0\% to 11.4\%) and a 34.4\% reduction in $\text{f}_\text{inc}$ (from 64.0\% to 42.0\%). This significant improvement validates VDM's effectiveness in addressing prior view biases. When combined with $\mathcal{L}_\text{P}$, the performance gains are further amplified. Our complete method (VDM + $\mathcal{L}_\text{P}$) achieves the best overall performance with $\text{f}_\text{mf}$ of only 3.4\% and $\text{f}_\text{inc}$ of 31.0\%, representing reductions of 96.0\% and 68.1\% compared to the baseline, respectively. Compared to Perp-Neg + $\mathcal{L}_\text{P}$, our method further reduces $\text{f}_\text{mf}$ by 77.8\% and $\text{f}_\text{inc}$ by 41.2\%, demonstrating the synergistic effects of VDM and $\mathcal{L}_\text{P}$ in effectively mitigating the multi-face Janus problem.

\begin{figure*}
    \centering
    \includegraphics[width=1\linewidth]{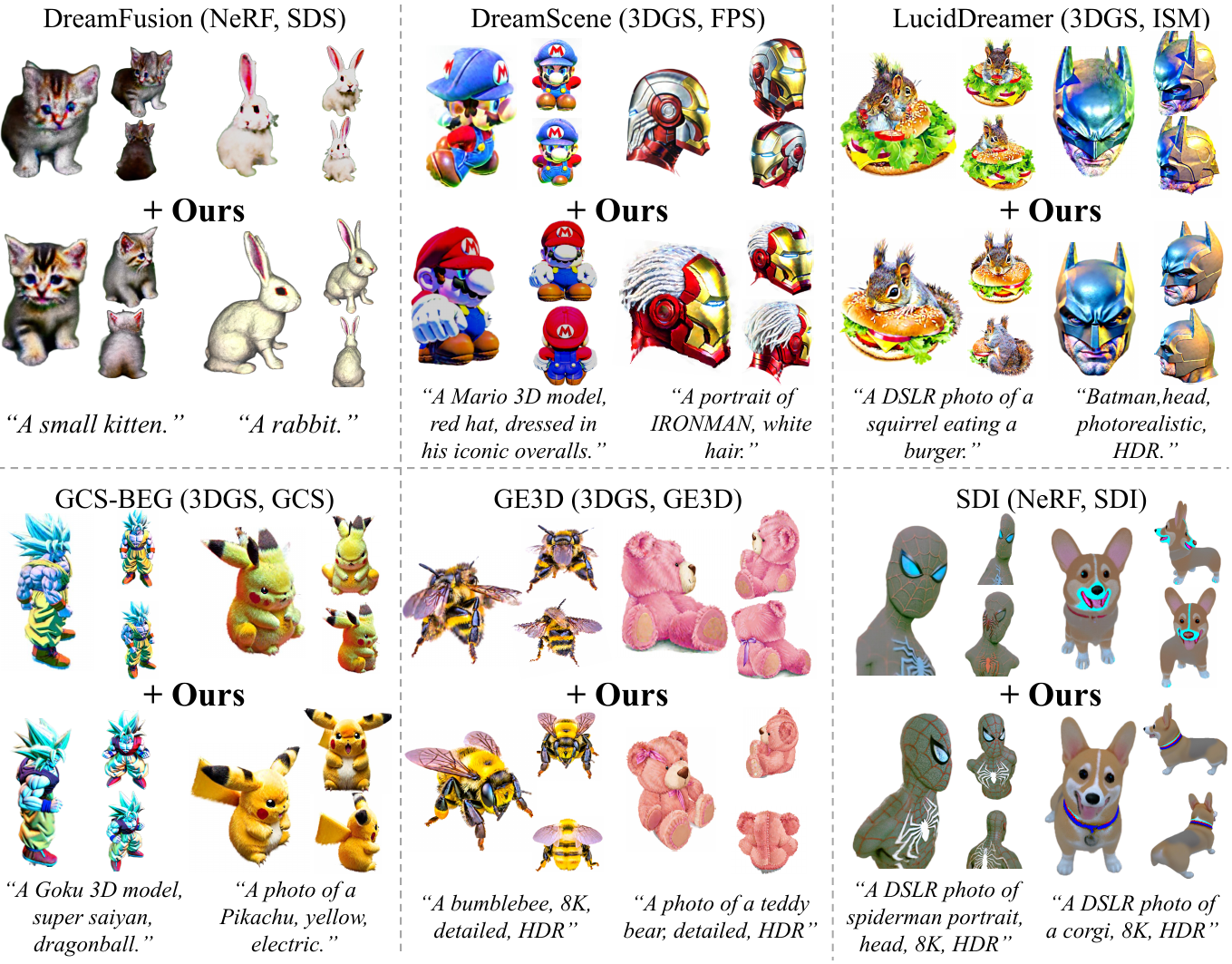}
    \caption{Qualitative text-to-3D generation results of our method. We integrate ConsDreamer into six SOTA baselines spanning NeRF/3DGS representations and diverse score distillation strategies, and these results demonstrate that our proposed method is visually superior to the baselines.}
    \label{Fig_12}
\end{figure*}

However, we chose those generation results with the multi-face Janus problem and observed that the OpenCLIP-L/14 metric yields higher scores in Table~\ref{Tab_2}. Fig.~\ref{Fig_11} presents the OpenCLIP similarity distribution between prompts and rendered results for different methods, including samples exhibiting the multi-face Janus problem for comparison. The similarity scores generally decrease progressively from the front view toward both ends, with the lowest scores observed for the back view, which aligns with intuition, as rendered results from the back view typically contain less information, leading to lower similarity between the prompt and the image for that view. However, when the generated object exhibits the multi-face Janus problem, features of prior preference views appear across multiple angles, resulting in abnormally high similarity scores for those views. This observation prompts us to explore more suitable ablation study methods. To more accurately evaluate the consistency of 3D-generated content, analysing the distribution of similarity scores across views is more effective than relying solely on the average similarity score. The results demonstrate that our method produces a similarity score distribution that aligns more closely with the ideal one.

\begin{table*}[t]
\centering
\large
\caption{Quantitative Results of Integrating ConsDreamer into Various Baselines}
\label{Tab_3}
\resizebox{\textwidth}{!}{
\begin{tabular}{lcccccccccccccc}
\toprule
\multirow{2}{*}{\textbf{Methods}} & \multirow{2}{*}{\begin{tabular}[c]{@{}c@{}}\textbf{3D}\\\textbf{Repr.}\end{tabular}} & \multirow{2}{*}{\begin{tabular}[c]{@{}c@{}}\textbf{Score}\\\textbf{Distill.}\end{tabular}} & \multirow{2}{*}{\begin{tabular}[c]{@{}c@{}}\textbf{Consistency}\\\textbf{Strategy}\end{tabular}} & \multicolumn{3}{c}{\textbf{User Study}} & \multirow{2}{*}{\textbf{ImageReward} $\uparrow$} & \multirow{2}{*}{\textbf{OpenCLIP} $\uparrow$} & \multicolumn{2}{c}{\textbf{Inconsist. Freq.}} & \multirow{2}{*}{\begin{tabular}[c]{@{}c@{}}\textbf{Time}\\\textbf{ (min)}\end{tabular}} & \multicolumn{2}{c}{\textbf{VRAM}} \\
\cmidrule(lr){5-7} \cmidrule(lr){10-11} \cmidrule(lr){13-14}
& & & & \textbf{Quality} $\uparrow$ & \textbf{Consistency} $\uparrow$ & \textbf{Avg} $\uparrow$ & & & \textbf{f$_{\text{mf}}$(\%)} $\downarrow$ & {\textbf{f$_\text{inc}$(\%)}} $\downarrow$ & & \textbf{Avg(GiB)} & \textbf{Peak(GiB)} \\
\midrule
DreamFusion & NeRF & SDS & \textbackslash & $3.48 \pm 1.25$ & $4.37 \pm 1.18$ & 3.93 & $-0.289 \pm 0.81$ & $0.307 \pm 0.025$ & 60.0 & 86.7 & $\sim$ 40 & 4.8 & 6.2 \\
\rowcolor{gray!15}
+ Ours & NeRF & SDS & ConsDreamer & $\textbf{5.45} \pm 1.02$& $\textbf{5.58} \pm 0.98$ & \textbf{5.52} & $\textbf{0.103} \pm 0.79$ & $\textbf{0.317} \pm 0.023$ & \textbf{26.7} & \textbf{60.0} & $\sim$ 42 & 4.9 & 6.3 \\

SDI & NeRF & SDI & Perp-Neg & $6.33  \pm 0.89$ & $6.89 \pm 0.76$& 6.61 & $0.297 \pm 0.61$ & $0.328 \pm 0.024$ & 33.3 & 61.9 & $\sim$ 119 & 28.0 & 39.2 \\
\rowcolor{gray!15}
+ Ours & NeRF & SDI & ConsDreamer & $\textbf{6.51} \pm 0.85$ & $\textbf{7.03} \pm 0.72$& \textbf{6.77} & $\textbf{0.175} \pm 0.59$ & $\mathbf{0.331} \pm 0.021$ & \textbf{27.0} & \textbf{40.5} & $\sim$ 116 & 27.8 & 39.0 \\

DreamScene & 3DGS & FPS & Perp-Neg & $6.25 \pm 0.92$ & $6.90 \pm 0.81$ & 6.58 & $0.201 \pm 0.75$ & $0.321 \pm 0.027$ & 48.3 & 66.7 & $\sim$ 30 & 9.8 & 13.6 \\
\rowcolor{gray!15}
+ Ours & 3DGS & FPS & ConsDreamer & $\textbf{6.93} \pm 0.88$ & $\textbf{7.45} \pm 0.75$ & \textbf{7.19} & $\mathbf{0.531} \pm 0.52 $ & $\textbf{0.324} \pm 0.027$ & \textbf{28.3} & \textbf{45.0} & $\sim$ 30 & 9.8 & 13.6 \\

LucidDreamer & 3DGS & ISM & Perp-Neg & $6.76 \pm 0.78$ & $6.31 \pm 0.95$ & 6.54 & $0.285 \pm 0.69$ & $0.328 \pm 0.019$ & 24.0 & 64.0 & $\sim$ 55 & 16.0 & 22.1 \\
\rowcolor{gray!15}
+ Ours & 3DGS & ISM & ConsDreamer & $\textbf{7.85} \pm 0.71$ & $\textbf{8.05} \pm 0.68$ & \textbf{7.95} & $\mathbf{0.707} \pm 0.41$ & $\mathbf{0.333} \pm 0.015$ & \textbf{3.4} & \textbf{31.0} & $\sim$ 55 & 16.0 & 22.0 \\

GCS-BEG & 3DGS & GCS & Perp-Neg & $7.03 \pm 0.84$ & $7.65 \pm 0.70$& 7.34 & $0.333 \pm 0.83$ & $0.316 \pm 0.031$ & 33.3 & 64.3 & $\sim$ 41 & 12.3 & 16.4 \\
\rowcolor{gray!15}
+ Ours & 3DGS & GCS & ConsDreamer & $\textbf{7.97} \pm 0.76$ & $\textbf{8.16} \pm 0.65$ & \textbf{8.07} & $\mathbf{0.814} \pm 0.51$ & $\textbf{0.327} \pm 0.035$ & \textbf{7.1} & \textbf{35.7} & $\sim$ 41 & 12.3 & 16.3 \\

GE3D & 3DGS & GE3D & Perp-Neg & $6.58 \pm 0.86$ & $6.93 \pm 0.79$ & 6.76 & $0.297 \pm 0.47$ & $0.324 \pm 0.020$ & 34.3 & 65.7 & $\sim$ 109 & 10.4 & 12.5 \\
\rowcolor{gray!15}
+ Ours & 3DGS & GE3D & ConsDreamer & $\textbf{7.01} \pm 0.82$ & $\textbf{7.64} \pm 0.73$ & \textbf{7.33} & $\mathbf{0.447} \pm 0.44$ & $\textbf{0.324} \pm 0.019$ & \textbf{17.1} & \textbf{36.7} & $\sim$ 109 & 10.3 & 12.5 \\
\bottomrule
\end{tabular}
}
\end{table*}

\subsection{\textit{Integration with SOTA Methods}}
\label{Comparison with SOTA Methods}

\textit{1) Qualitative Results:} Fig.~\ref{Fig_2} presents vivid and diverse text-to-3D generation results produced by our method. Furthermore, we integrate ConsDreamer into a range of advanced 3D generation frameworks, which cover different 3D representation paradigms and various score distillation strategies. As demonstrated in Fig.~\ref{Fig_12}, our method substantially alleviates the multi-face Janus issue prevalent in existing approaches.

\begin{figure*}
    \centering
    \includegraphics[width=0.96\linewidth]{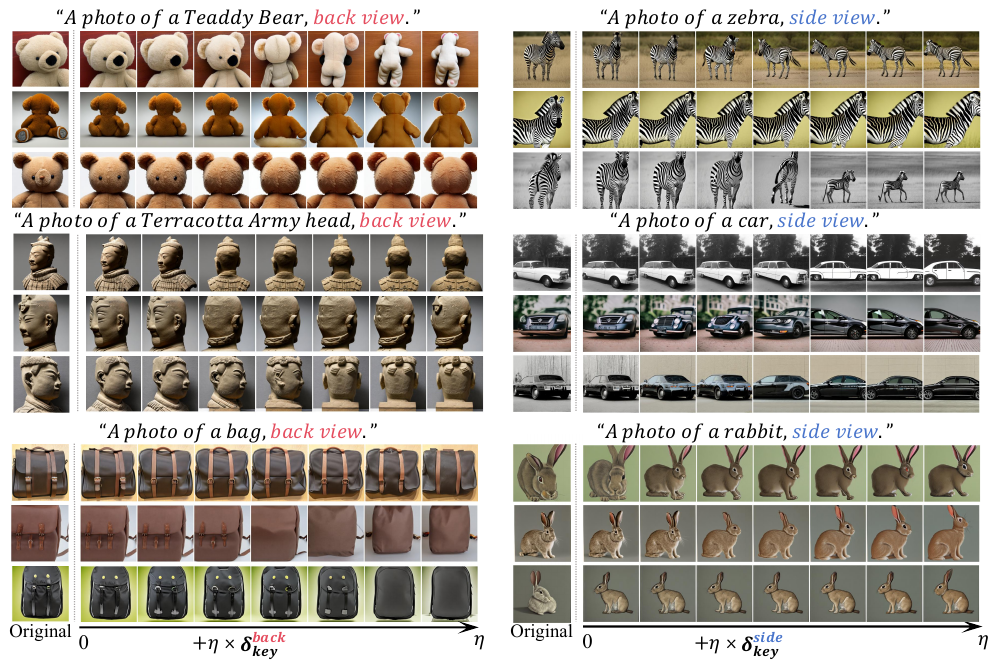}
    \caption{ Application of the VDM to T2I Tasks. The results demonstrate the outputs of Stable Diffusion with explicit view descriptions across six prompts. The first column in each set shows the original outputs, highlighting the inherent view bias in T2I models. By applying the VDM, the view features were disentangled, enabling precise control of desired views (e.g., ``back view" or ``side view").}
    \label{Fig_13}
\end{figure*}

\textit{2) Quantitative Results:}  We selected ImageReward, OpenCLIP-L/14, and inconsistency frequency as objective evaluation metrics. As shown in Table~\ref {Tab_3}, ConsDreamer delivers more substantial gains over the baseline than PerpNeg, attaining state-of-the-art performance across all metrics. Specifically, in the multi-face Janus issue rate, our method outperforms PerpNeg-based baselines by approximately 52.3\%. Additionally, we report the mean runtime and GPU VRAM usage. For GPU VRAM usage, we compute the average of the maximum GPU memory consumption across multiple runs. Table~\ref{Tab_3} demonstrates that our method is lightweight, as it does not compromise GPU memory or runtime for consistent generation.

We conducted a user study involving 145 participants to subjectively compare our method with baselines in terms of the quality and view consistency of the generated 3D content. Approximately 75\% of participants were undergraduate and graduate students majoring in AI-related fields, 20\% were students from non-AI disciplines, and the remaining 5\% were non-student participants. The questionnaire included detailed representative cases and scoring guidelines to standardise the evaluation. Participants rated each model on a 1–10 scale across two key dimensions: ``Generation Quality" and ``3D Consistency". To ensure fairness, all samples were randomly selected without cherry-picking to favor our method. As shown in Table~\ref{Tab_3}, our method achieves the highest average scores across both evaluation metrics compared to baselines. These results demonstrate that the 3D content generated by our approach is consistently more appealing and exhibits superior view consistency.

\subsection{\textit{Application of the VDM to T2I Tasks}}
\label{Application of the VDM}
\textit{1) Qualitative Results:}
To further validate the effectiveness of the VDM, we extended its application to 2D generation tasks. As shown in Fig.~\ref{Fig_13}, we present 18 sets of generation results from the Stable Diffusion model across six prompts, each containing explicit view descriptions (e.g., ``back view" or ``side view"). The first column in each set shows the original outputs, which reveal the inherent view bias in T2I models. By directly applying the VDM, the view features are effectively disentangled, enabling explicit control over the desired views. The results clearly demonstrate that the VDM mitigates view semantic errors and significantly improves generation accuracy.

\textit{2) Quantitative Results:} We compared the success rate of generating images with specific views using Stable Diffusion, Perp-Neg and our method. Given the open-ended nature of the generated content, a strict definition was adopted for counting success cases. Images exhibiting hallucinations (e.g., artefacts or incorrect semantics) were excluded. Additionally, in instances where multiple entities appeared in an image, it was considered a failure if even one entity did not match the specified view. The results, presented in Table~\ref{Tab_4}, show that our proposed VDM improves rear-view generation success by approximately 36.7\% over Perp-Neg, demonstrating that our method achieves a substantially higher success rate and effectively aligns multimodal view semantics.

\begin{table}
\centering
\begin{threeparttable} 
\caption{Success Rate of Specific View Generation}
\label{Tab_4}
\begin{tabular}{l c c}
\toprule
\multirow{2}{*}{\textbf{Methods}} & \multicolumn{2}{c}{ {\textbf{Successful Generation Rate}}{\textbf{(\%)}}$ \uparrow$} \\
\cmidrule(lr){2-3}
& \textbf{Side view} & \textbf{Back view} \\
\midrule
Stable Diffusion & 66.3 & 28.7 \\
Stable Diffusion + Perp-Neg & 77.4 & 56.5 \\
Stable Diffusion + VDM (Ours) & \textbf{80.2} & \textbf{77.2} \\
\bottomrule
\end{tabular}
\end{threeparttable}
\end{table}

\begin{figure}
    \centering
    \includegraphics[width=1\linewidth]{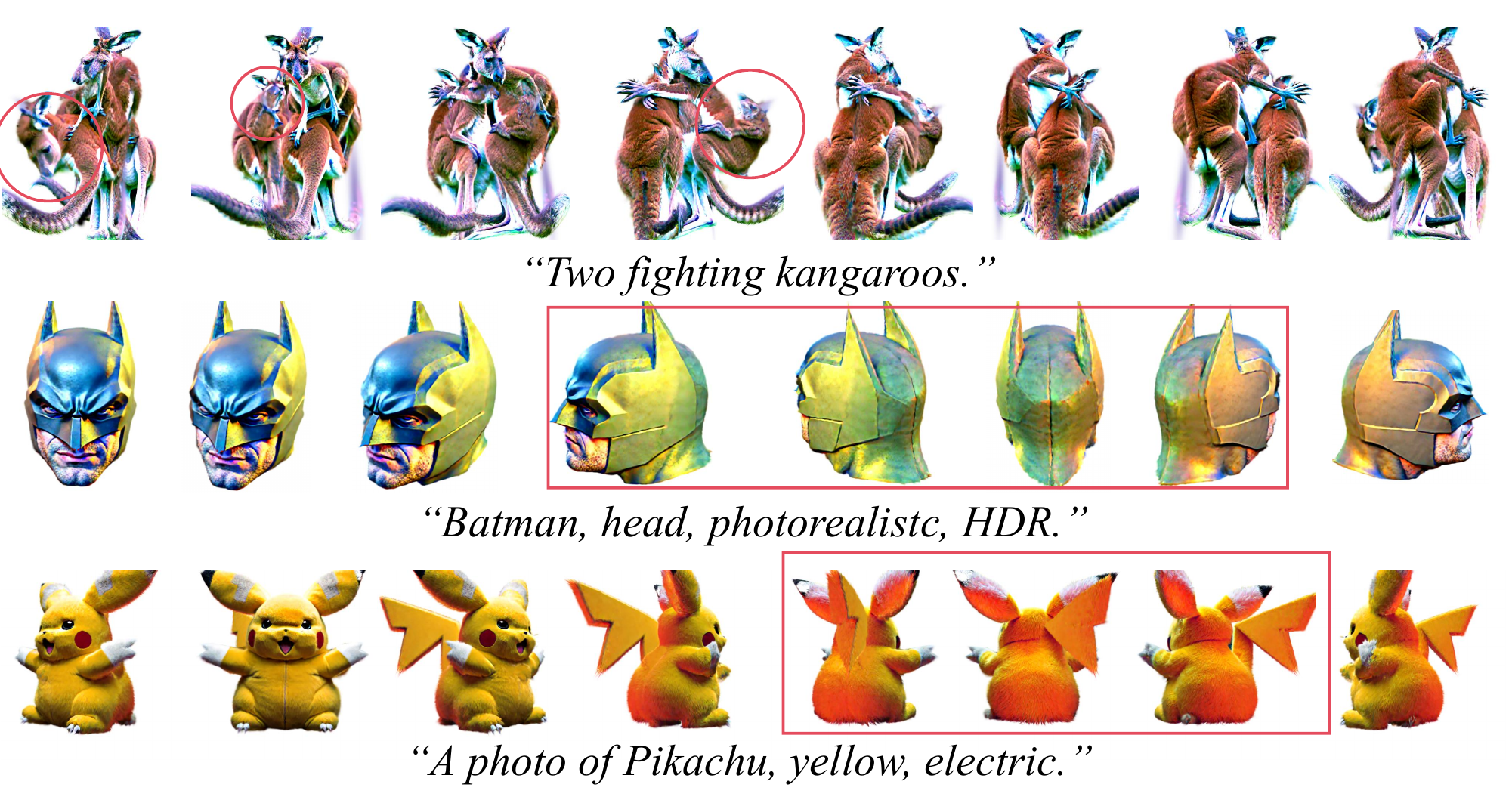}
    \caption{Generation failure case examples. For complex or multi-object prompts, the underlying optimisation may still produce distorted or entangled geometry (top, red circles), while for single-object cases, ConsDreamer improves view consistency but cannot fully remove illumination artefacts such as oversaturation and colour shifts (middle and bottom, red boxes).}
    \label{Fig_14}
\end{figure}

\subsection{\textit{Limitations}}
\label{Application of the VDM}
Methods built on score-distillation pipelines, while capable of substantially improving multi-view consistency, still inherit the fundamental limitations of this optimisation paradigm. For highly complex text descriptions or cluttered multi-object scenes, the underlying optimisation may converge to sub-optimal 3D configurations, leading to fragmented or entangled structures. In these regimes, partial-order constraints act only as a relatively weak regularisation signal and cannot by themselves guarantee globally coherent geometry, as illustrated in Fig.~\ref{Fig_14}. Moreover, score-distillation-based approaches predominantly target view-level semantics and geometry, and typically do not explicitly account for illumination artefacts introduced during distillation. Brightness accumulation and colour saturation anomalies can therefore persist even when view consistency is improved (Fig.~\ref{Fig_14}). A principled treatment of such photometric effects would likely require revisiting the distillation objective itself, which we regard as an important direction for future work.

\section{Conclusion} In this paper, we conducted a mathematical analysis of the multi-face Janus problem in zero-shot text-to-3D generation. While Previous methods alleviate this problem by orthogonalizing negative prompts in the denoising score space, they only operate on the UNet output and overlook the aggregated prior view features formed during the UNet processing, leading to the recurrence of inconsistency when lifting 2D representations to 3D. To address this fundamental limitation, we propose ConsDreamer, which offers two key advantages: (1) a View Disentanglement Module that eliminates view biases in conditional prompts by decoupling irrelevant view components and injecting precise view control prior to denoising; and (2) a similarity-based partial order loss that enforces geometric consistency in the unconditional term by aligning cosine similarities with azimuthal relationships, synergistically enhancing view semantic clarity. Demonstrating strong versatility, ConsDreamer can be seamlessly integrated into diverse 3D representations (e.g., NeRF, 3DGS) and various score distillation paradigms. Extensive experiments across multiple SOTA baselines confirm that our method effectively mitigates the multi-face Janus problem while outperforming existing approaches in both visual quality and multi-view consistency.

\bibliographystyle{IEEEtran}
\bibliography{main}

\begin{IEEEbiography}[{\includegraphics[width=1in,height=1.25in,clip,keepaspectratio]{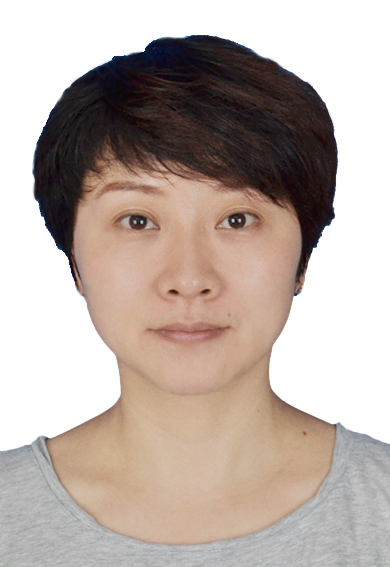}}]{Yuan Zhou}
received the Ph.D. degree from Nanjing University of Aeronautics and Astronautics, Nanjing, China, in 2016. She was an Academic Visitor with the University of East Anglia, Norwich, U.K., from 2017 to 2018. She is currently an Associate Professor with the School of Artificial Intelligence, Nanjing University of Information Science and Technology, Nanjing, China. Her research interests include zero-shot learning, multimodal fusion, and computer vision.\end{IEEEbiography}

\begin{IEEEbiography}[{\includegraphics[width=1in,height=1.25in,clip,keepaspectratio]{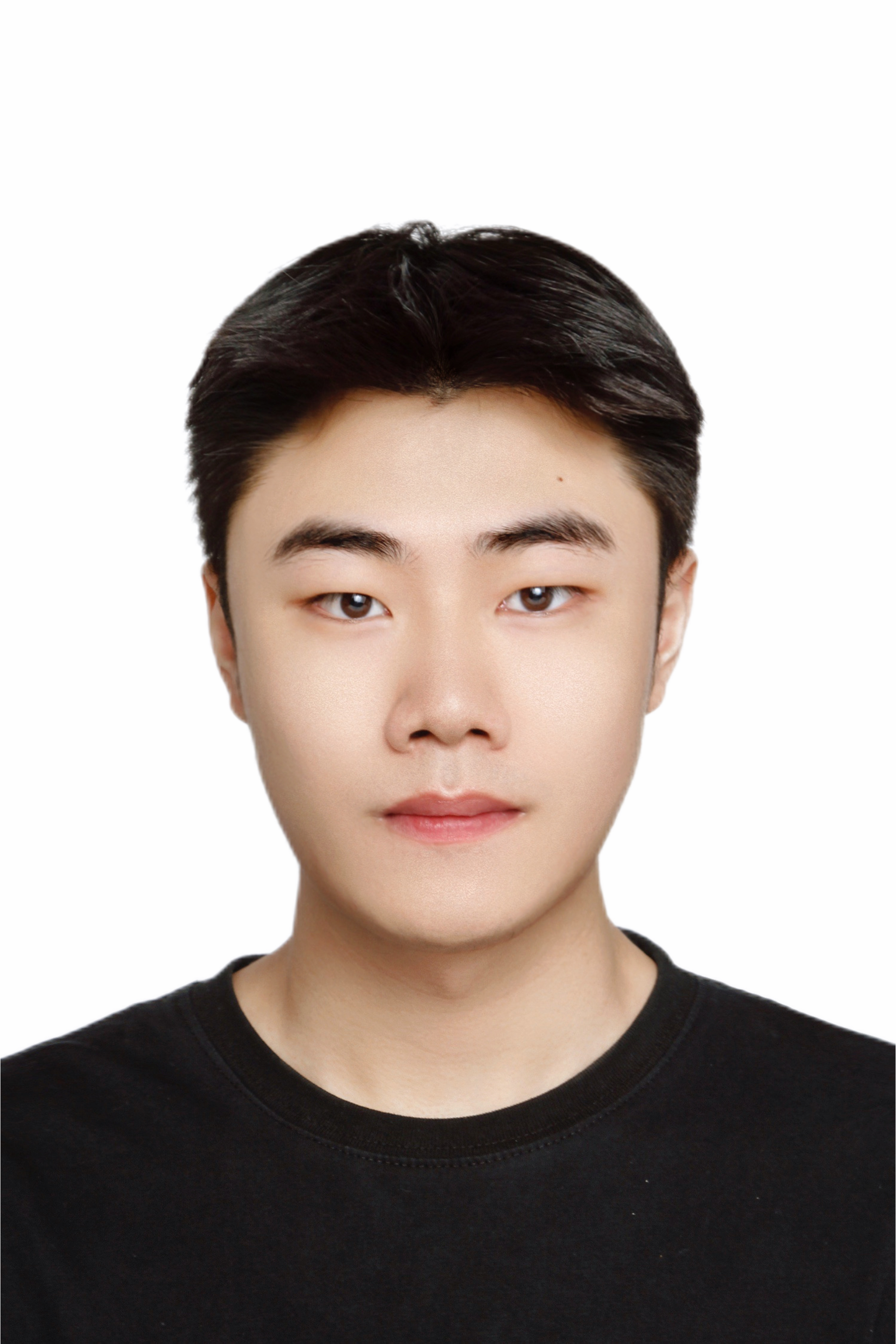}}]{Shilong Jin} is currently pursuing the M.E. degree in artificial intelligence at Nanjing University of Information Science and Technology, Nanjing, China. He will pursue the Ph.D. degree with the School of Artificial Intelligence, Nanjing University of Aeronautics and Astronautics, Nanjing, China. He previously interned at Lenovo Research, Beijing, China, with the CTOO Human-centered Innovation Intelligence and Insights department in 2025. His research interests include 3D generation and editing, embodied intelligence, and pose estimation.\end{IEEEbiography}

\begin{IEEEbiography}[{\includegraphics[width=1in,height=1.25in,clip,keepaspectratio]{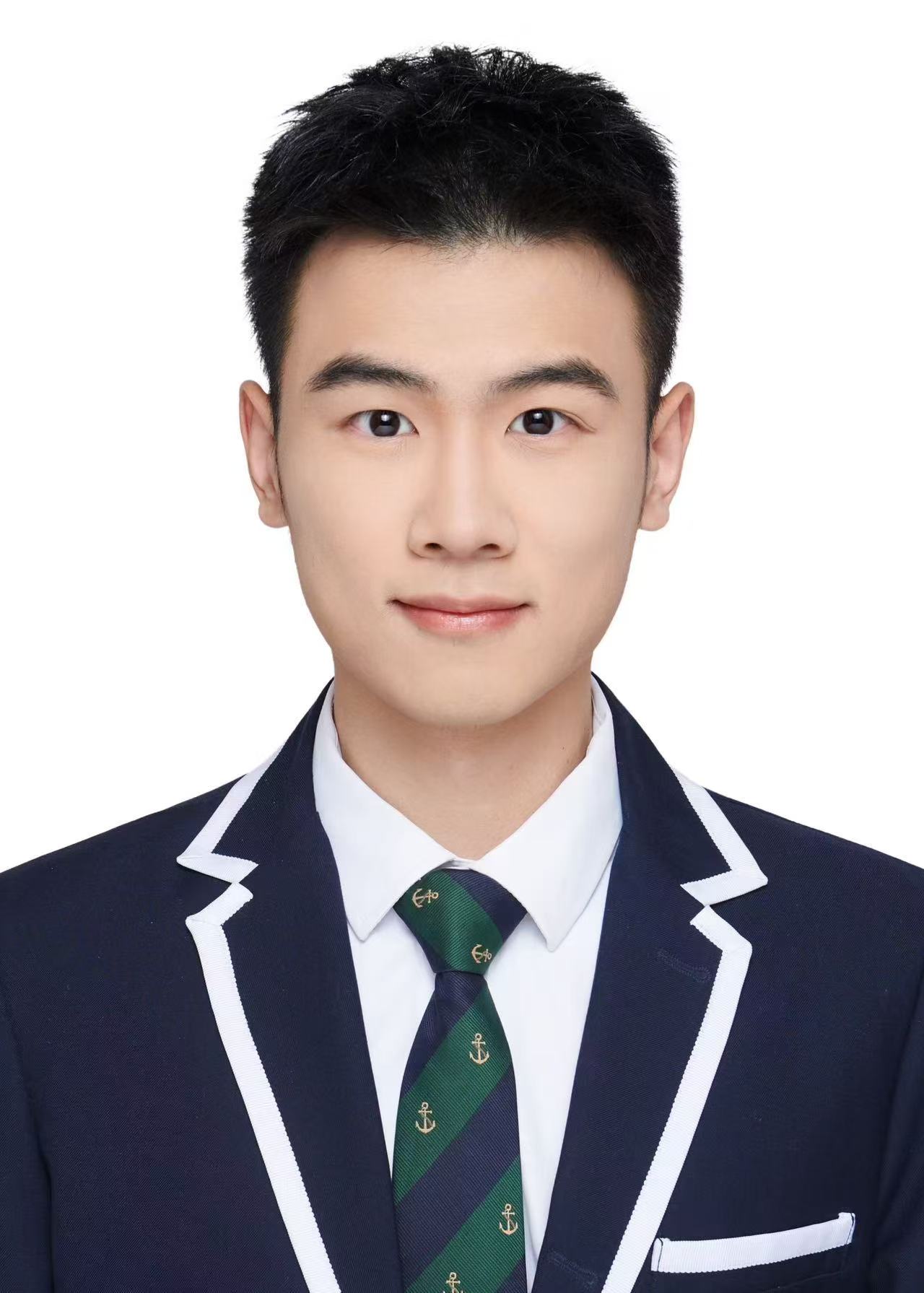}}]{Litao Hua}
received the B.E. degree in Artificial Intelligence from Nanjing Univeristy of Information Science and Technology, Nanjing, China, in 2023. He is currently pursuing the M.E. degree in Artificial Intelligence with Nanjing University of Information Science and Technology, Nanjing, China. His research interests include 3D vision, video understanding and Embodied Intelligence.\end{IEEEbiography}

\begin{IEEEbiography}[{\includegraphics[width=1in,height=1.25in,clip,keepaspectratio]{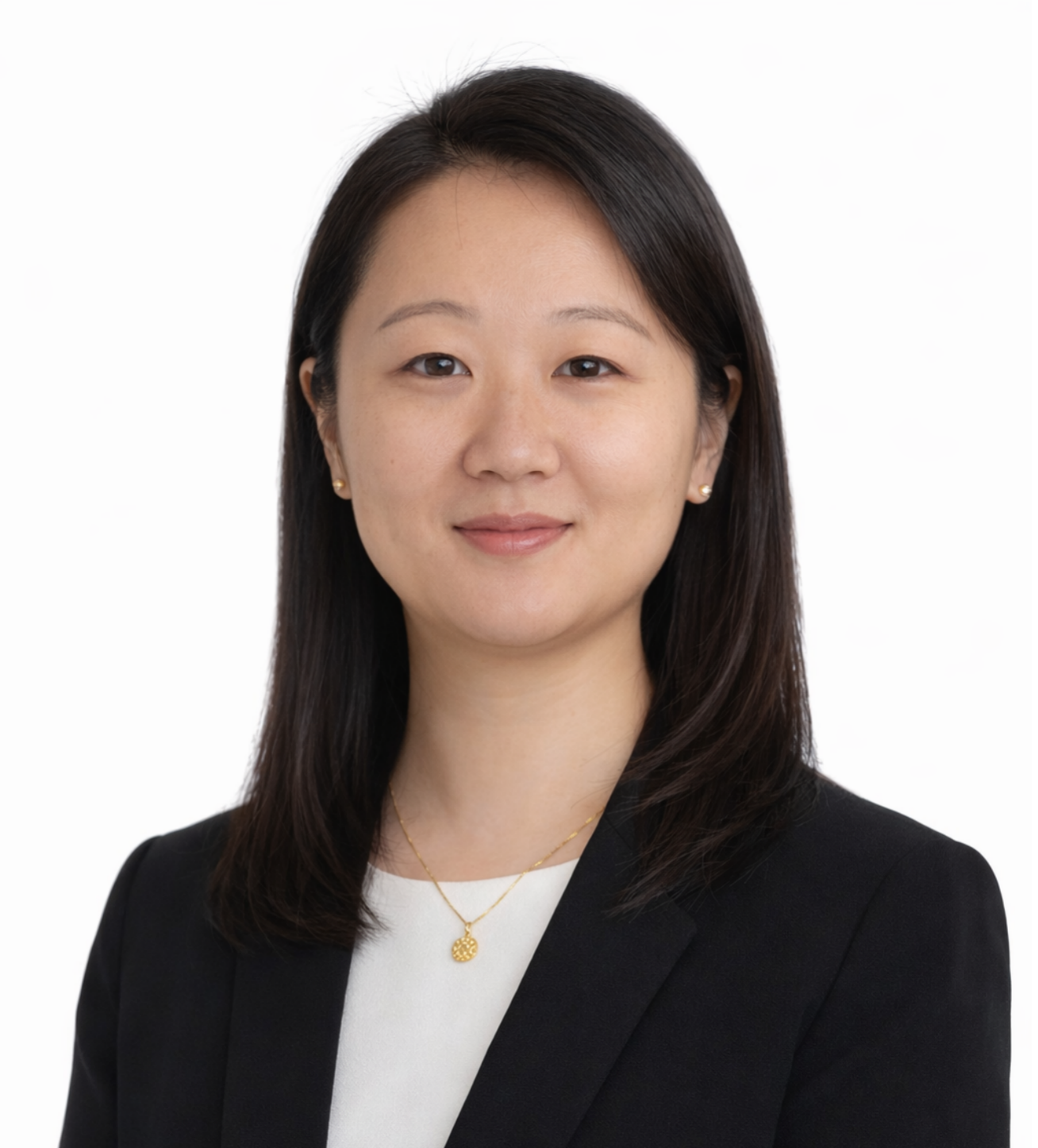}}]{Wanjun Lv} received the B.E. degree in communication engineering and the M.E. degree in electronic and communication engineering from Xi’an Jiaotong University, Xi’an, China, in 2011 and 2013, respectively. She has been with Lenovo CTO Org since 2013, and is currently a Senior Researcher. Her research interests include computer vision, 3D vision and multimodal human–agent interaction. She has worked on glass-free 3D spatial interaction, 3D content generation, OCR, and handwriting recognition. She holds more than 21 granted patents.\end{IEEEbiography}

\begin{IEEEbiography}[{\includegraphics[width=1in,height=1.25in,clip,keepaspectratio]{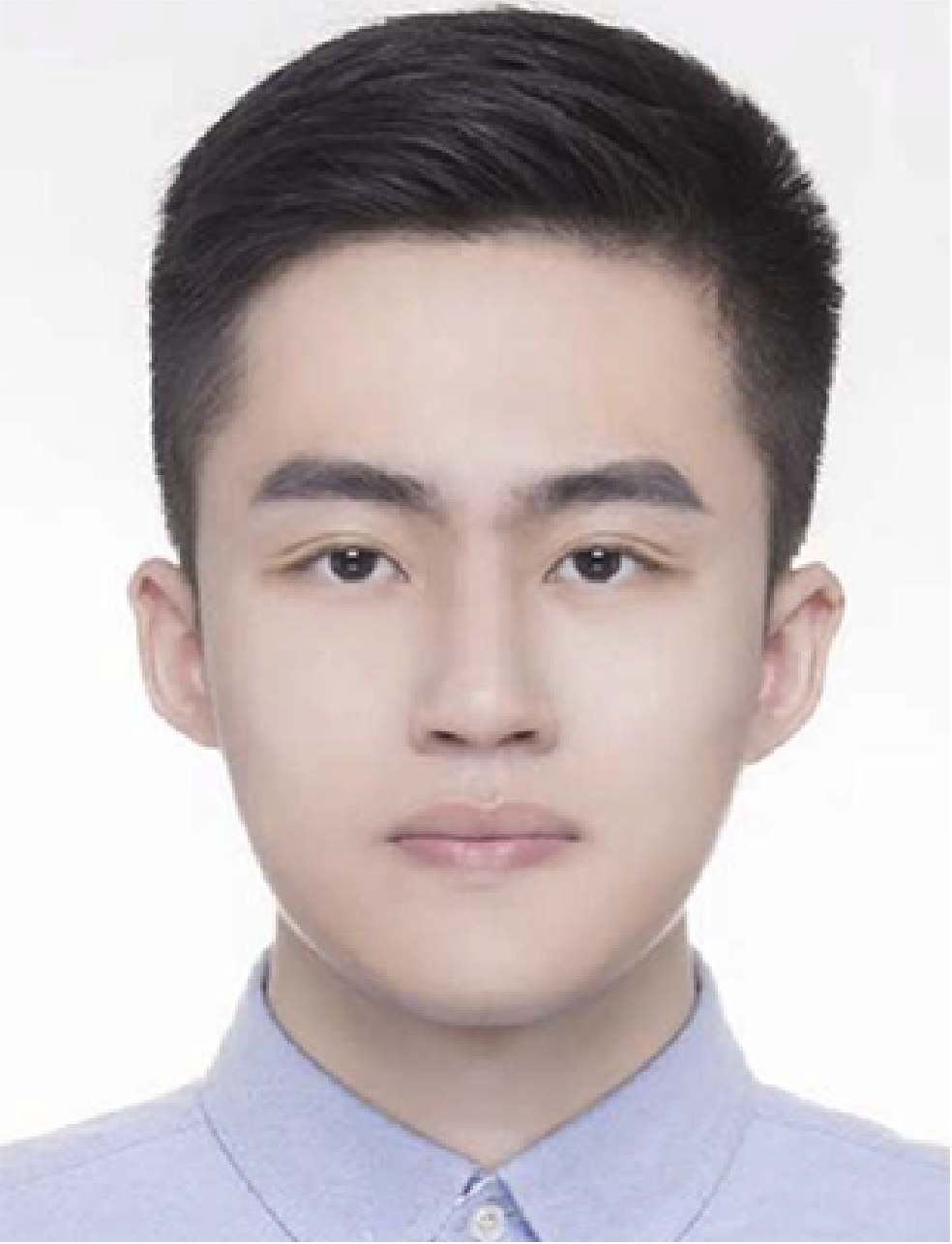}}]{Haoran Duan} (Member, IEEE) received the M.S. degree (Hons.) in data science from Newcastle University, Newcastle upon Tyne, U.K., in 2019, and the Ph.D. degree in artificial intelligence from the Department of Computer Science, Durham University, Durham, U.K. He was a Research Assistant with OpenLab, Newcastle University, and a Research Associate with the School of Computing, Newcastle University, where he worked on deep learning applications. He is currently a Postdoctoral Research Associate with the Department of Automation, Tsinghua University, Beijing, China. His research interests include the theory and applications of deep learning and multimodal computer vision.\end{IEEEbiography}

\begin{IEEEbiography}[{\includegraphics[width=1in,height=1.25in,clip,keepaspectratio]{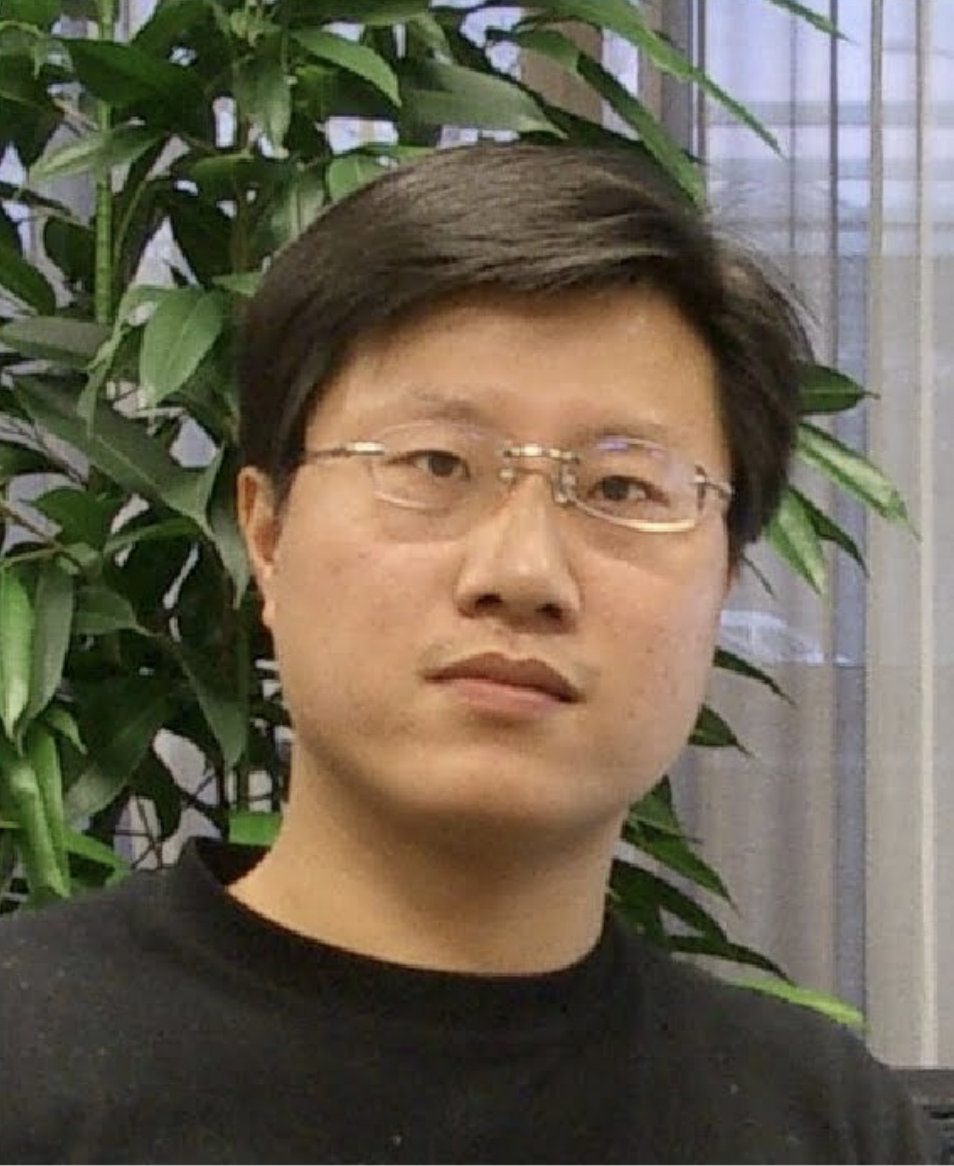}}]{Jungong Han} (Senior Member, IEEE) received the Ph.D. degree in telecommunication and information system from Xidian University, Xi’an, China, in 2004. He is currently a Professor with the Department of Automation, Tsinghua University, Beijing, China. He has published more than 200 articles, including more than 80 IEEE Transactions and more than 50 A* conference papers. His research interests include video analysis, computer vision, and applied machine learning. He is a fellow of the International Association of Pattern Recognition.\end{IEEEbiography}
\end{document}